\documentclass[journal]{IEEEtran}
\usepackage{amsmath,amsfonts}
\usepackage{algorithmic}
\usepackage{algorithm}
\usepackage{array}
\usepackage{booktabs}
\usepackage[caption=false,font=normalsize,labelfont=sf,textfont=sf]{subfig}
\usepackage{textcomp}
\usepackage{multirow}
\usepackage{color} 
\usepackage[normalem]{ulem}
\useunder{\uline}{\ul}{}
\usepackage{stfloats}
\usepackage{url}
\usepackage{verbatim}
\usepackage{graphicx}
\usepackage{cite}
\usepackage{hyperref}
\usepackage{setspace}
\usepackage[mathscr]{euscript}
\begin{document}
\title{SFTformer: A Spatial-Frequency-Temporal Correlation-Decoupling Transformer for Radar Echo
Extrapolation}

 \author{Liangyu~Xu, Wanxuan~Lu, Hongfeng~Yu, {Fanglong~Yao}, Xian Sun, \IEEEmembership{Senior~Member,~IEEE,} Kun~Fu, \IEEEmembership{Senior~Member,~IEEE}% <-this % stops a space
\thanks{\textit{(Corresponding authors: Wanxuan~Lu, Hongfeng~Yu)}}

\thanks{~Liangyu~Xu, ~Xian Sun,~Kun Fu are with the Aerospace Information Research Institute, Chinese Academy of Sciences, Beijing 100094, China, also with the Key Laboratory of Network Information System Technology (NIST), Aerospace Information Research Institute, Chinese Academy of Sciences, Beijing 100190, China, also with the University of Chinese Academy of Sciences, Beijing 100049, China, and also with the School of Electronic, Electrical and Communication Engineering, University of Chinese Academy of Sciences, Beijing 100049, China (e-mail: xuliangyu21@mails.ucas.ac.cn; sunxian@aircas.ac.cn; kunfuiecas@gmail.com).}%
\thanks{~Wanxuan~Lu,~Hongfeng Yu and ~Fanglong~Yao are with the Aerospace Information Research Institute, Chinese Academy of Sciences, Beijing 100094, China, and also with the Key Laboratory of Network Information System Technology (NIST), Aerospace Information Research Institute, Chinese Academy of Sciences, Beijing 100190, China (e-mail: luwx@aircas.ac.cn; yuhf@aircas.ac.cn; yaofanglong17@mails.ucas.ac.cn).}
}

\maketitle

\begin{abstract}
Extrapolating future weather radar echoes from past observations is a complex task vital for precipitation nowcasting. The spatial morphology and temporal evolution of radar echoes exhibit a certain degree of correlation, yet they also possess independent characteristics. {Existing methods learn unified spatial and temporal representations in a highly coupled feature space, emphasizing the correlation between spatial and temporal features but neglecting the explicit modeling of their independent characteristics, which may result in mutual interference between them.} To effectively model the spatiotemporal dynamics of radar echoes, we propose a Spatial-Frequency-Temporal
correlation-decoupling Transformer (SFTformer). The model leverages stacked multiple SFT-Blocks to not only mine the correlation of the spatiotemporal dynamics of echo cells but also avoid the mutual interference between the temporal modeling and the spatial morphology refinement by decoupling them. Furthermore, inspired by the practice that weather forecast experts effectively review historical echo evolution to make accurate predictions, SFTfomer incorporates a joint training paradigm for historical echo sequence reconstruction and future echo sequence prediction. Experimental results on the HKO-7 dataset and ChinaNorth-2021 dataset demonstrate the superior performance of SFTfomer in short(1h), mid(2h), and long-term(3h) precipitation nowcasting. 
\end{abstract}

\begin{IEEEkeywords}
 Correlation-decoupling, SFT-Block, Prediction-Reconstruction Joint Training Paradigm
\end{IEEEkeywords}

\section{Introduction}
\begin{figure}
\centering
\includegraphics[width=1.0\linewidth]{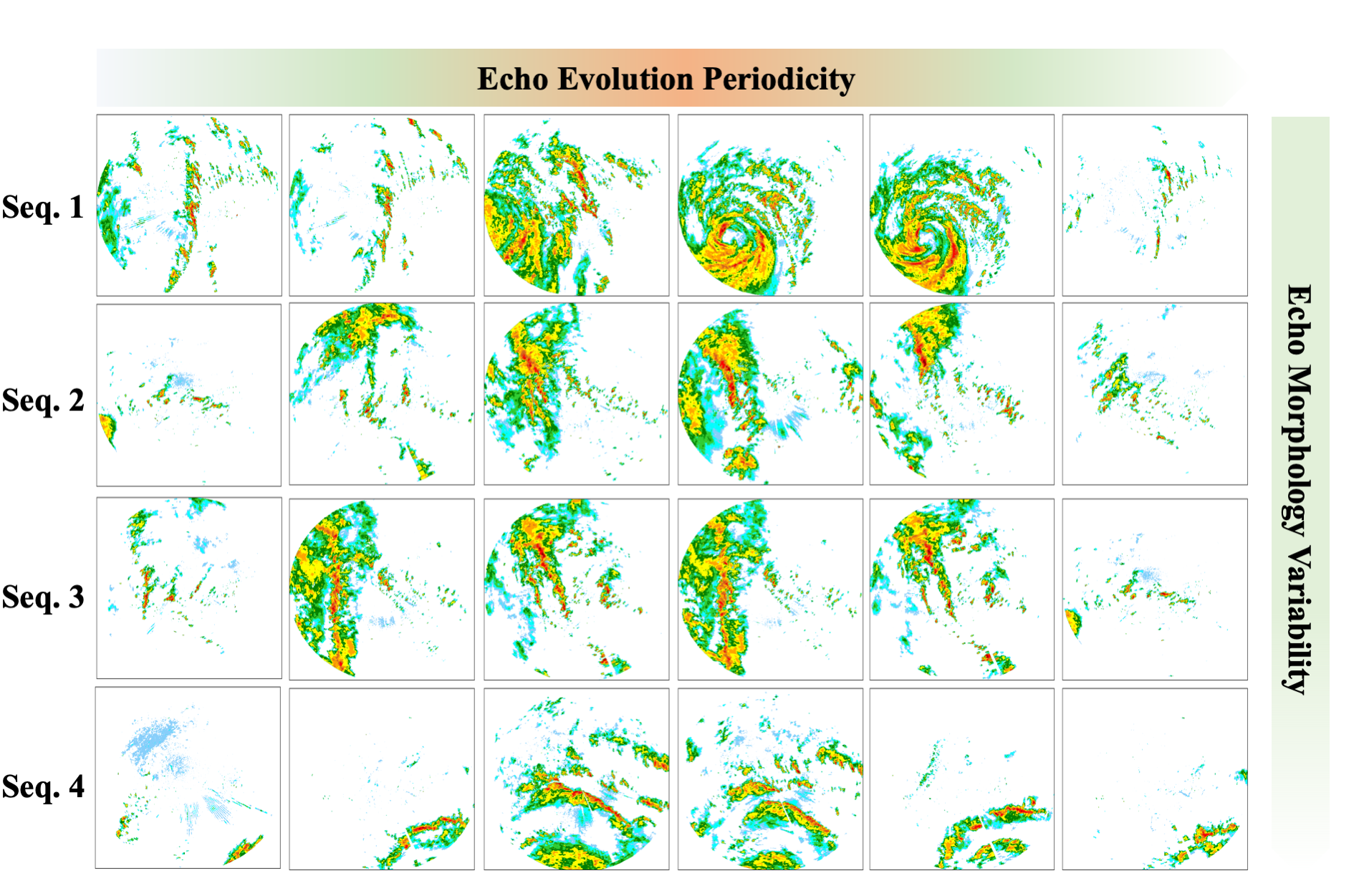}
\caption{Radar echo sequences observed during 4 periods from HKO-7 dataset. The 4 sequences exhibit distinct spatial morphologies but share a common evolution lifecycle characterized by initiation,
maturation and decay of echo evolution.}
\label{fig_intro}
\end{figure}

\IEEEPARstart{R}{adar} echo extrapolation involves utilizing the past radar echo data to predict future radar echoes, which plays a crucial role in delivering accurate precipitation nowcasting and timely warnings for catastrophic weather events, thereby ensuring vital information dissemination to the community. {The core challenge of radar echo extrapolation lies in reasoning the future motion of radar echoes (temporal evolution) while clearly modeling their appearance (spatial morphology). In fact, there is a certain correlation between the motion of echoes and their appearance, while they also possess independence. The correlation is evident when the spatial distribution of echoes exhibits specific appearance, such as a circular or curved shape, typically indicating that it will undergo some form of motion in the future, such as rotation or displacement. The independence of the motion of echoes from their appearance is manifested in different weather systems, although the morphology (appearance) of echoes varies, they still undergo common motion (temporal evolution) features, such as aggregation and dissipation.}  In the four radar echo sequences from HKO-7 shown in Fig. \ref{fig_intro}, despite their different spatial distributions, they all undergo a process from initiation to maturation and eventually decay, which represents the temporal evolution characteristics that are relatively independent of the spatial morphology.

Manifesting the chaotic nature of atmospheric evolution, radar echo extrapolation techniques face considerable challenges in modeling the intricate dynamics of the atmosphere. Traditional radar echo extrapolation methods rely on computing motion fields between adjacent echo frames and linearly extrapolating the results to forecast precipitation. However, these methods are only suitable for cases with smooth and simple motion patterns over short periods. Typical methods include optical flow\cite{woo2017operational,ayzel2019optical}, centroid tracking\cite{dixon1993titan,johnson1998storm,del2018radar}, and cross-correlation\cite{li1995nowcasting,mecklenburg2000improving,liang2010composite}. Over the past few years, with the notable progress and extensive utilization of deep learning, radar echo extrapolation has been increasingly regarded as spatiotemporal predictive learning\cite{babaeizadeh2017stochastic, wang2019memory, srivastava2015unsupervised, wang2018eidetic, weissenborn2019scaling, rakhimov2020latent,ranzato2014video,kalchbrenner2017video}. As a result, researchers have embraced deep learning methods\cite{DBLP:journals/corr/ShiCWYWW15,DBLP:journals/corr/abs-2103-09504,wang2019memory,shi2017deep,wu2021motionrnn}, particularly variants of recurrent neural networks (RNN\cite{medsker2001recurrent}s), to enhance the accuracy and efficacy of this technique. The RNN-based approaches organize radar echo data over time steps to form input sequences. At each time step, the model receives the current radar echo data and integrates past time step information, iteratively updating its internal hidden state. The iterative process enables the model to capture dynamic patterns and trends within the time series, thereby predicting future radar echo conditions.  

However, two prominent problems still exist. On one hand, these models {typically handle the prediction of radar echo motion and appearance modeling within the same set of modules, learning a unified spatiotemporal representation of echoes in a highly coupled feature space. Although the correlation between echo appearance and motion is taken into account, these methods often involve complex spatiotemporal feature extraction and feature transfer operations, leading to inevitable mutual interference.} On the other hand, such models still face inherent limitations of recurrent neural network architectures. In radar echo extrapolation, distant history information forgetting is a challenge where models may not effectively retain and utilize information from earlier time steps when dealing with long sequences, which occurs due to the inherent characteristics of recurrent neural networks (RNNs). RNNs process sequences by iteratively updating their hidden states, which can cause information from distant past time steps to gradually fade away. The challenge of distant history information forgetting arises from several factors, including the vanishing gradient problem, limited memory capacity, and the iterative nature of updating hidden states. Over time, as the sequence length increases, the ability of the model to remember and use information from the distant past can diminish. Although some variants\cite{wang2019memory,DBLP:journals/corr/abs-2103-09504,DBLP:journals/corr/abs-1804-06300} have mitigated the impact of this problem through delicate architectural design, the fundamental issue remains unsolved due to the limitations of the architecture.

To address the issues mentioned above, we deviate from the prevalent use of RNN-based architecture in existing research and instead utilize the potent feature extraction capabilities of the Transformer\cite{DBLP:journals/corr/VaswaniSPUJGKP17}. Adhering to the idea of decoupling spatial morphology and temporal evolution, we propose a Spatial-Frequency-Temporal correlation-decoupling Transformer (SFTformer). Specifically, SFTformer consists of stacked SFT-Blocks, aiming to tackle the challenges in representation learning caused by strong spatiotemporal coupling. The SFT-Block mines correlations between spatial morphology and temporal evolution while modeling their individual characteristics. Within each block, a hierarchical architecture of correlation-decoupling is employed. The spatiotemporal correlation layer roughly models the global spatiotemporal correlation, while the spatial refinement layer refines the morphological features of the echo cells, and the temporal modeling layer mines the temporal evolution. Notably, the temporal modeling layer not only provides equal attention to each time step, addressing the issue of forgetting historical information, but also models the cyclic pattern of echo evolution through frequency analysis. Moreover, to enhance the memory of historical information, a joint training paradigm is proposed, involving the reconstruction of historical echo sequences and the prediction of future echo sequences. By reconstructing the features of historical echo sequences while making predictions, the model generates more accurate forecasts. In addition, leveraging the richly learned radar echo evolution characteristics, the model directly predicts all future frames, effectively avoiding the accumulation of errors. The primary contributions of the research are outlined below:
\begin{itemize}

\item{We propose a novel Transformer-based radar echo extrapolation method named SFTformer, whose core component SFTBlock adopts hierarchical correlation-decoupling architecture to explore the spatiotemporal correlations of echoes while finely modeling the spatial morphology and temporal evolution.
}

\item{Inspired by the practice of weather forecast experts who review historical echo evolutionary patterns to make a weather forecast, we design a joint training paradigm for historical echo sequence reconstruction and future echo sequence prediction. By consolidating the memory of historical information, the model can provide more accurate radar echo predictions.}

\item{Through experimental evaluations on both the HKO-7 dataset and ChinaNorth-2021 dataset, our model demonstrates state-of-the-art performance in short-term (1-hour), mid-term (2-hour), and long-term (3-hour) precipitation forecasting.}
\end{itemize}

The remaining sections of this paper are organized as follows: In Section \ref{sec:related}, we review spatiotemporal predictive learning and the application of Transformer in sequence prediction and computer vision domains. Section \ref{sec_method} presents the problem definition and our proposed SFTformer. The experimental results and analysis are detailed in Section \ref{sec:experiment}. Finally, we conclude the paper in Section \ref{sec:conclusion}.

\section{Related Works} \label{sec:related}
In this section, a brief review of classical spatiotemporal prediction methods is presented, encompassing radar echo extrapolation techniques and video prediction methods. Additionally, the application of Transformer in time series prediction and computer vision tasks is introduced.
 
 \subsection{Deep Learning-based Radar Echo Extrapolation}
 Over the past few years, with the notable progress and extensive utilization of deep learning\cite{bi2023not,mao2022beyond,mao2023elevation,yao2023automated,yao2023ringmo,DBLP:journals/corr/VaswaniSPUJGKP17}, researchers have embraced deep learning methods to enhance the accuracy and efficacy of radar echo extrapolation. ConvLSTM\cite{shi2015convolutional} augments the traditional fully connected LSTM by integrating convolutional structures, enabling it to effectively capture spatiotemporal correlations.There are also some methods specifically designed for radar echo extrapolation tasks. TrajGRU\cite{shi2017deep} models natural motion and rotation transformations to learn the position changes in recursive connections and proposes a comprehensive evaluation scheme for future research to measure the state of the art. HPRNN\cite{jing2020hprnn} caters to the necessity of long-term extrapolation in practical forecasting by implementing hierarchical prediction strategies and a recurrent coarse-to-fine mechanism, which significantly reduces the accumulation of prediction errors over time. REMNet\cite{jing2022remnet} introduces the Long-term Evolution Rule Memory (LERM) module to effectively model the lifecycle of actual echo sequences from initiation to decay. {Unlike these methods, our method departs from the conventional use of RNN and adopts the Transformer architecture. Moreover, we design a correlation-decoupling hierarchical structure to learn the spatiotemporal dynamics of echoes.}
 
\subsection{Spatiotemporal Predictive Learning}
 The mainstream approach for spatiotemporal predictive learning adopts RNN-based architectures to learn the dynamics of spatiotemporal sequences. PredRNN\cite{wang2017predrnn} puts forward a framework that enables the simultaneous extraction and memorization of both temporal and spatial representations of sequences. PredRNN++\cite{wang2018predrnn++} aims to address the challenges of gradient propagation and long-term dependency capture by incorporating the highway unit. PredRNNv2\cite{wang2022predrnn} expands PredRNN by decoupling the loss and utilizing the reverse scheduled sampling method. MIM\cite{wang2019memory} acknowledges the presence of both stationary and non-stationary characteristics in spatiotemporal sequences and introduces the self-renewed memory module to effectively model these properties in videos. Due to their excellent flexibility and prediction accuracy, these methods play a fundamental role in spatiotemporal predictive learning. {In comparison, our method is specifically designed for radar echo extrapolation, considering the correlation and independence of appearance and motion information in radar echoes.}
 
\subsection{Transformers for Time Series Forecasting and Vision Tasks}
Transformer\cite{vaswani2017attention} is a neural network that relies on a self-attention mechanism. It diverges from traditional convolutional neural networks and recurrent neural networks by employing a fully connected attention mechanism to model and predict sequences, offering parallel computing capability and the capacity to model long-term dependencies. While Transformer has not been explored in radar echo extrapolation research, it has demonstrated impressive achievements in time series prediction and computer vision domains. In the realm of time series prediction, Transformer has been introduced to capture long-term dependencies and has shown impressive performance\cite{zhou2021informer,wu2021autoformer,beltagy2020longformer,so2019evolved,hua2022transformer}. For further improving the long-term prediction capabilities of Transformer, FEDformer\cite{zhou2022fedformer} utilizes the sparse representation characteristics of many time series in known bases like Fourier transform and develops frequency-enhanced Transformer. In computer vision, ViT\cite{dosovitskiy2020image} has made groundbreaking progress by directly applying the Transformer architecture to image classification on non-overlapping patches of medium-sized images, achieving an impressive balance between speed and accuracy. To address the differences between text and images, Swin Transformer\cite{liu2021swin} proposes a hierarchical structure. The representation is computed using a shifting window scheme, which restricts the self-attention calculation to non-overlapping local windows while still permitting cross-window connections. This design enhances efficiency during computation. {Although these methods have not been directly applied to radar echo extrapolation, they demonstrate the powerful capabilities of Transformers in time sequence modeling and provide valuable references for introducing Transformer into radar echo extrapolation. The proposed method adapts the attention computation method to suit the characteristics of radar echoes.}
\section{Methodology} \label{sec_method}
This section provides a detailed explanation of the proposed method. Section \ref{sec_method:pre} presents the formulation of radar echo extrapolation and Section \ref{sec_method:overall} describes the overall framework of the network. Section \ref{sec_method:SFT} describes in detail the SFT-Block, consisting of the spatiotemporal correlation layer, spatial refinement layer and temporal modeling layer. Section \ref{sec_method:Restruction} introduces the prediction-reconstruction joint training paradigm, focusing on describing the reconstruction methods, as well as the loss function of joint training. 

\subsection{Preliminary} \label{sec_method:pre}
Suppose we observe a sequence of radar echo ${\mathcal{X}}_{t,T}={\{x_i\}}_{t-T+1}^{t}$ of length T at a fixed time interval, where {$x_i\in \mathbb{R}^{1 \times H \times W}$} is the radar echo at time step $i$. The input channel is 1, representing the logarithmic radar reflectivity factor (unit: dBZ), which serves as an indicator of precipitation particle scale. H and W denote the spatial height and width, respectively. The ground truth of the forecasting result can be expressed as ${\mathcal{Y}}_{t+1, T^{'}}={\{y_i}\}_{t+1}^{t+T^{'}}$.
The model $\operatorname{F}$ with the learnable parameter $\boldsymbol{\theta} $ learns the mapping from ${\mathcal{X}}_{t,T}$ to ${\mathcal{Y}}_{t+1, T'}$, i.e., $\operatorname{F_\theta}:  {\mathcal{X}}_{t,T} \mapsto {\mathcal{Y}}_{t+1,T'}$. By fully utilizing the typical characteristics of echo sequences, such as their spatiotemporal correlations and motion patterns, our main objective is to minimize the difference between the predicted sequence $\operatorname{F_\theta}({\mathcal{X}}_{t, T})$ and the ground truth ${\mathcal{Y}}_{t+1, T'}$. Furthermore, meteorological experts often enhance future forecasting accuracy by reviewing historical patterns of echo evolution. To simulate this process, we design a module to reconstruct the spatial features of historical echoes. The training objective of the model is to simultaneously minimize both prediction errors and reconstruction errors.

\begin{equation}
    \boldsymbol{\theta}^{\ast}=\mathop{\arg\min}\limits_{\theta}\left(\mathcal{L}(\operatorname{F_{\theta}}({\mathcal{X}}_{t,T}),{\mathcal{Y}}_{t+1,T^{'}}) + \lambda \mathcal{L}_{recon}\right)
    \label{optim_Eq}
\end{equation}

where $\boldsymbol{\theta}^{\ast}$ denotes the model parameters capable of minimizing both the prediction loss and the reconstruction loss, $\mathcal{L}$ represents the loss function employed to quantify the disparity between the forecasting result and the ground truth, $\mathcal{L}_{recon}$ represents the reconstruction loss, and $\lambda$ is the weight controlling the two losses. The detailed calculation process of $\mathcal{L}_{recon}$  is described in Section \ref{sec_method:Restruction}.

\subsection{Overall Framework} \label{sec_method:overall}
\begin{figure*}
\centering
\includegraphics[width=1.0\linewidth]{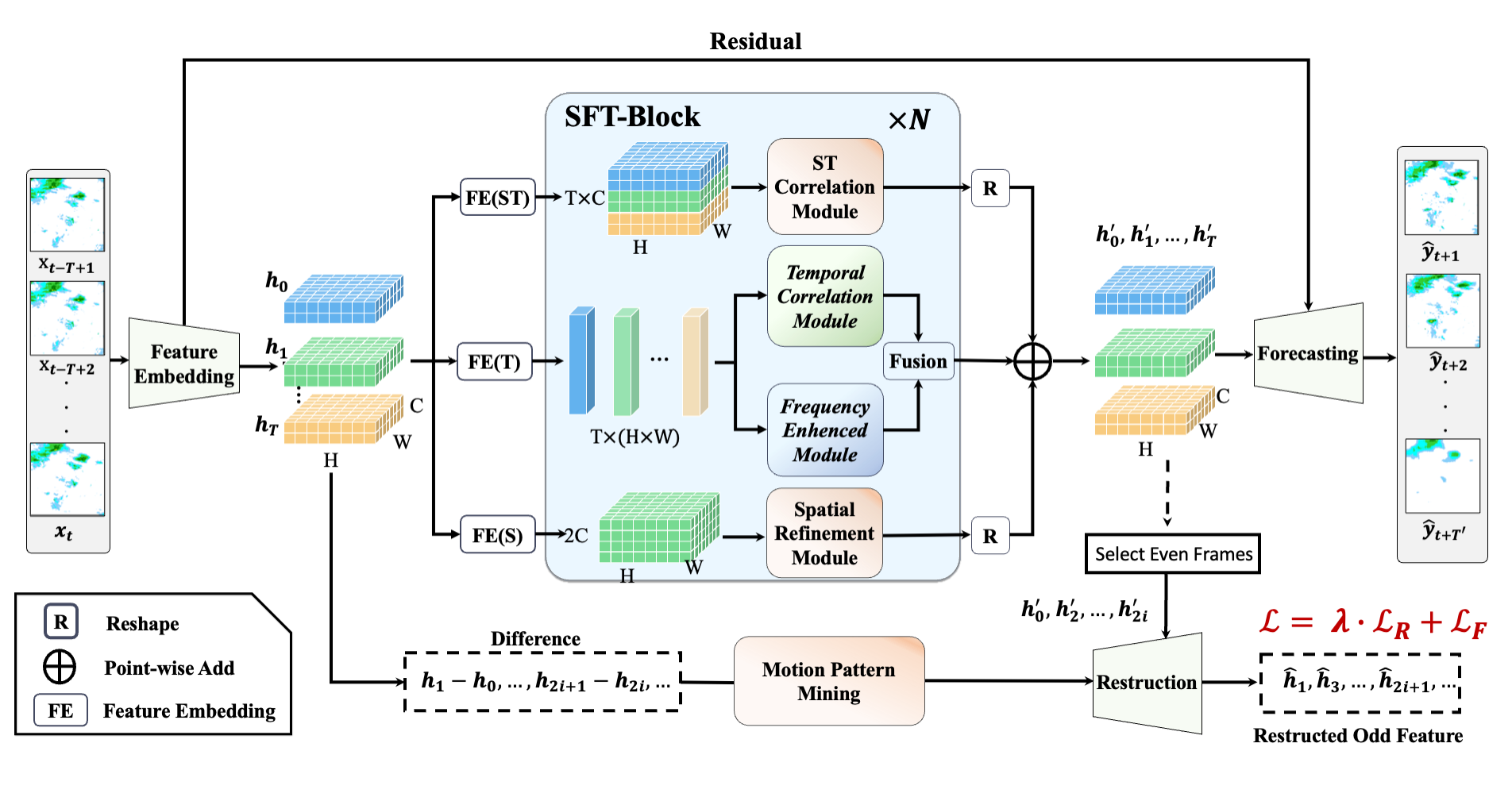}
\caption{The overall framework of SFTformer. The historical echo sequence ${\mathcal{X}}_{t, T}$ is compactly embedded to ${\mathcal{H}}$ through Feature Embedding and then entered into N stacked SFT-Blocks to learn spatiotemporal dynamics.  SFT-Block has a correlation-decoupling hierarchical architecture. The spatiotemporal correlation layer mines global coarse spatiotemporal correlation. The spatial refinement module models the refined morphological features of radar echo using a higher channel dimension. The temporal modeling layer learns temporal evolution patterns by means of temporal interaction and frequency analysis. The features after SFT-Block ${\mathcal{H}}^{'}$ pass through the Forecasting module (spatial upsampling) to the final prediction $\boldsymbol{\hat{Y}}$. In addition, the reconstruction branch excavates the motion pattern of the echo, and the even-frame features in ${\mathcal{H}}^{'}$ jointly recover the odd-frame features in H through the Reconstruction module.}
\label{fig_overall_architecture}
\end{figure*}

SFTformer is an implementation of $F_{\theta}$ for radar echo extrapolation mentioned in Section \ref{sec_method:pre}, and the overall framework is illustrated in Fig. \ref{fig_overall_architecture}. The historical echo sequences are first passed through the Feature Embedding ($\operatorname{FE}$) module to obtain more compact feature representations. The FE module preserves the original temporal dimension of the echo sequences while operating exclusively on the spatial dimension, compressing the original echo sequences to obtain the implicit spatial representation ${\mathcal{H}}=\{h_i\}_{t-T+1}^t$, where $h_i$ represents the feature of the i-th echo image in the historical echo sequences, situated within the implicit spatial space. Specifically, the process of obtaining $h_i$ can be represented as:
\begin{equation}
     h_{i}^{j} = \operatorname{\sigma}\left(\operatorname{GN}\left(\operatorname{Conv2d}\left(h_{i}^{j-1}\right)\right)\right)
\end{equation}
where $ 1 \leq j \leq M $, $\sigma$ represents the non-linear activation function, GN refers to GroupNorm2d, i represents the index of the time series, and j represents the j-th layer of the FE module. Specifically, $h_i^0 = x_i$, and $h_i = h_i^M$. When j is an even number, the Conv2d stride is configured to 2 for downsampling, while for odd-numbered layers, it is set to 1.\par

To effectively capture the dynamics of historical radar echoes,  ${\mathcal{H}}$ is initially fed into stacked SFT-Blocks to learn the spatiotemporal evolution of radar echoes. The formulation can be expressed as follows:
\begin{equation}
    {\mathcal{H}}^{k} = \operatorname{SFT-Block}({\mathcal{H}}^{k-1})
\end{equation}
Where $M<k\leq M+N$. Each SFT-Block adopts a correlation-decoupling hierarchical strategy, which enables the extraction of both coarse spatiotemporal coupling correlation and the decoupled learning of spatial morphology (spatial-domain features), and the evolution patterns (temporal-domain features). The specific details will be elaborated in Section \ref{sec_method:SFT}.\par
The output ${\mathcal{H}}^{M+N}$ from the last layer of the SFT-Block is fed into the Forecasting module to output the final prediction results. The specific process is as follows:

\begin{equation}
    {\mathcal{H}}^{l} = \operatorname{\sigma}\left(\operatorname{GN}\left(\operatorname{unConv2d}\left({\mathcal{H}}^{l-1}\right)\right)\right)
\end{equation}

Where $M+N < l < 2M+N$,  ${\mathcal{H}}^{2M+N} = \operatorname{\sigma}\left(\operatorname{GN}\left(\operatorname{unConv2d}\left({\mathcal{H}}^{2M+N-1}+{\mathcal{H}}^{2}\right)\right)\right)$, unconv2d represents the transpose convolution operation. When $l$ is an even number, the stride of unconv2d is set to 2 for upsampling; otherwise, it is set to 1. The prediction result is represented as $\hat{{\mathcal{Y}}}={\mathcal{H}}^{2M+N}$.\par

Simultaneously, the implicit spatial features of the radar echoes denoted as ${\mathcal{H}}$, are also inputted into the reconstruction branch. By taking the difference between odd frames and even frames, the motion features of the echoes, denoted as D, are obtained. The Motion Pattern Mining module is utilized to capture the motion patterns of the echoes. Subsequently, together with the even frames from the output ${\mathcal{H}}^{M+N}$ of the last SFT-Block, the odd frames in the original features ${\mathcal{H}}$ are reconstructed using the Reconstruction module. By reconstructing the hidden spatial features, the model consolidates the memory of the historical echo sequences, thereby enhancing the predictive ability for future frames. The detailed implementation is described in Section \ref{sec_method:Restruction}. \par

\subsection{SFT-Block} \label{sec_method:SFT}
\begin{figure*}
\centering
\includegraphics[width=1.0\linewidth]{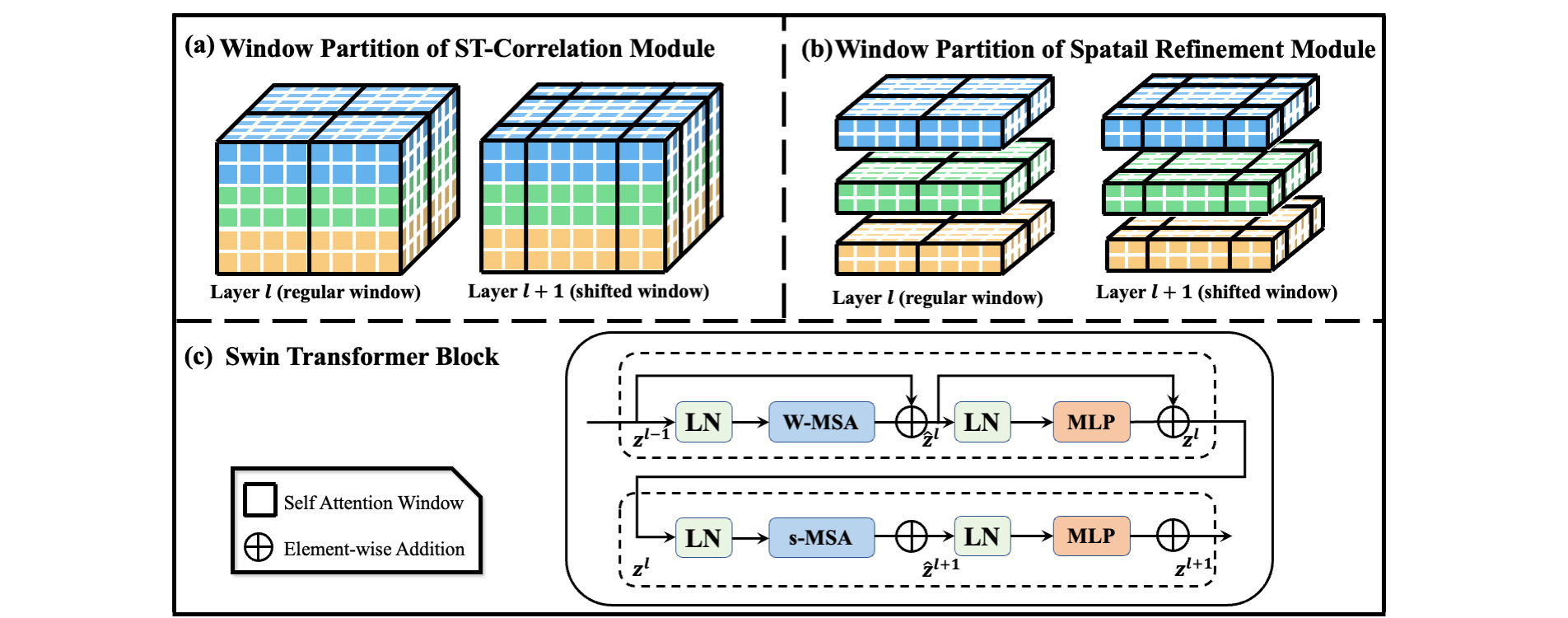}
\caption{Both the spatiotemporal correlation layer and the spatial refinement layer use the Swin Transformer block as the basic unit, but the window partitioning strategies for calculating attention are different: (a) for the ST-Correlation layer and (b) for the spatial refinement layer. (c) shows details of the Swin Block.}
\label{fig_swin}
\end{figure*}
In order to better capture the spatiotemporal evolution patterns of radar echoes, the SFT-Block is designed with a correlation-decoupling hierarchical structure. The spatiotemporal correlation layer, as shown in Fig \ref{fig_swin}(a), learns the global spatiotemporal correlations by stacking multiple frames of radar echo features along the time axis. By enforcing the spatiotemporal correlation module, built upon the Swin Transformer Block, to learn the stacked multi-frame features, the model becomes capable of capturing both the intrinsic temporal evolution and spatial morphological changes of radar echo sequences. As a result, the representations obtained are both spatially and temporally correlated.

To better handle the interference between spatial details and temporal evolution in radar echoes, we explicitly decouple the spatiotemporal characteristics of echoes. The spatial refinement layer models the spatial morphological details and the temporal modeling layer mines the temporal evolution patterns of echo cells. The temporal modeling layer effectively alleviates the issue of historical information forgetting by incorporating information exchange among multiple frames. Additionally, the temporal modeling layer includes frequency analysis to capture the periodic evolution patterns of the radar echoes. {Next, we will provide a detailed explanation of each layer. It is important to note that all the Feature Embedding modules mentioned below are executed before the SFT-Block.}

\subsubsection{Spatiotemporal Correlation Layer}
The spatiotemporal coupling correlation features refer to the representation of radar echo's spatial morphology and temporal evolution patterns within the same feature space. The representation encompasses not only the spatial characteristics and the temporal evolution patterns but also implies their mutual influence. {The apperance (spatial morphology) of radar echoes varies with the motion (temporal evolution) of echoes, indicating a correlation between them. One the one hand, when the spatial distribution of echoes exhibits a specific form, such as a circular or curved shape, it usually indicates that it will undergo some form of motion in the future, such as rotation or movement. On the other hand, when the echoes show an expanding motion trend, it often suggests an increase in the spatial distribution range. Therefore, studying the relationship between temporal evolution and spatial morphology of radar echoes can provide more comprehensive support for precipitation forecasting.}

The input of the spatiotemporal correlation layer, denoted as $z_{st} \in \mathbb{R}^{(T \times C) \times H \times W}$, is obtained from the initial features $z \in \mathbb{R}^{T \times C \times H \times W}$ through feature embedding (ST). The specific implementation involves stacking the multi-frame features along the time axis, which can be expressed as follows:
\begin{equation}
    z_{st} = [z_1;z_2;...;z_T]
\end{equation}
Then, $z_{st}$ are processed by the Swin Transformer Block. The Swin Transformer utilizes window-based MSA (W-MSA) and shifted window-based MSA (SW-MSA), which deviates from the conventional window partitioning approach and employs a more efficient shifted window partitioning strategy, as illustrated in Fig. \ref{fig_swin}(a). This approach not only maintains efficient computations with non-overlapping windows but also establishes connections between windows through shifting, as shown in Fig. \ref{fig_swin}(c). The expression can be summarized as follows:
\begin{equation}
    \hat{z}_{st}^{'}=\mathrm{W}-\mathrm{MSA}\left(\mathrm{LN}\left(z_{st}\right)\right)+z_{st} 
\end{equation}
\begin{equation}
z_{st}^{'}=\operatorname{MLP}\left(\mathrm{LN}\left(\hat{z}_{st}^{'}\right)\right)+\hat{z}_{st}^{'}
\end{equation}
\begin{equation}
     \hat{z}^{''}_{st}=\operatorname{SW}-\mathrm{MSA}\left(\mathrm{LN}\left(z_{st}^{'}\right)\right)+z_{st}^{'}
\end{equation}
\begin{equation}
z^{''}_{st}=\operatorname{MLP}\left(\mathrm{LN}\left(\hat{z}^{''}_{st}\right)\right)+\hat{z}^{''}_{st}
\end{equation}
where LN stands for Layer Normalization, (S)W-MSA represents (Shifted) Window-Multihead Self-Attention, and $z_{st}^{''} \in \mathbb{R}^{(T \times C) \times H \times W}$ denotes the output spatiotemporal correlation features, which are obtained through Reshape to yield the final output of this layer, $f_{st} \in \mathbb{R}^{T \times C \times H \times W}$. By compelling the spatiotemporal correlation layer to learn from stacked multi-frame features, the spatiotemporal correlation layer simultaneously captures the intrinsic temporal evolution patterns and spatial variations in radar echo sequences, resulting in coupled features that reflect the spatiotemporal correlations of the observation sequence.
\subsubsection{Spatial Refinement Layer}
Refining spatial features is crucial for learning radar echo spatiotemporal evolution patterns. Different precipitation types correspond to distinct radar echo distribution characteristics, which reflect the diverse attributes of weather systems. Furthermore, spatial feature refinement can also reveal the evolution patterns of weather systems. As weather systems develop and evolve, the distribution characteristics of radar echoes during different stages also undergo continuous changes. Thus, refining the spatial features of radar echoes can provide a more precise characterization of the evolution process and features of weather systems, improving the accuracy and reliability of weather forecasting.

The initial feature tensor, denoted as \(z \in \mathbb{R}^{T \times C \times H \times W}\), undergoes processing through FE(S) to produce the input for the spatial refinement layer. FE(S) is a 1$\times$1 convolution operation that doubles the number of channels in \(z\), resulting in \(z_s \in \mathbb{R}^{T \times 2C \times H \times W}\).  Having more channels allows the model to learn various types of features, aiding in capturing different characteristics and patterns in the input data. Subsequently, by using the Swin Transformer Block shown in Fig. \ref{fig_swin}(c), the layer learns rich representations of the radar echo morphology. The attention computation is performed within each frame, and the window partitioning method is illustrated in Fig. \ref{fig_swin}(b). The computation process can be expressed using the following formula:
\begin{equation}
\hat{z}_{s}^{'}=\mathrm{W}-\mathrm{MSA}\left(\mathrm{LN}\left(z_{s}\right)\right)+z_{s}
\end{equation}
\begin{equation}
z_{s}^{'}=\operatorname{MLP}\left(\mathrm{LN}\left(\hat{z}_{s}^{'}\right)\right)+\hat{z}_{s}^{'}
\end{equation}
\begin{equation}
    \hat{z}^{''}_{s}=\operatorname{SW}-\mathrm{MSA}\left(\mathrm{LN}\left(z_{s}^{'}\right)\right)+z_{s}^{'}
\end{equation}
\begin{equation}
z^{''}_{s}=\operatorname{MLP}\left(\mathrm{LN}\left(\hat{z}^{''}_{s}\right)\right)+\hat{z}^{''}_{s}
\end{equation}
where LN stands for Layer Normalization, and (S)W-MSA refers to (Shifted) Window-Multihead Self Attention. The output spatial refinement feature is denoted as $f_s=z_s^{''} \in \mathbb{R}^{T \times C \times H \times W}$.

\subsubsection{Temporal Modeling Layer}
\begin{figure}
\centering
\includegraphics[width=1.0\linewidth]{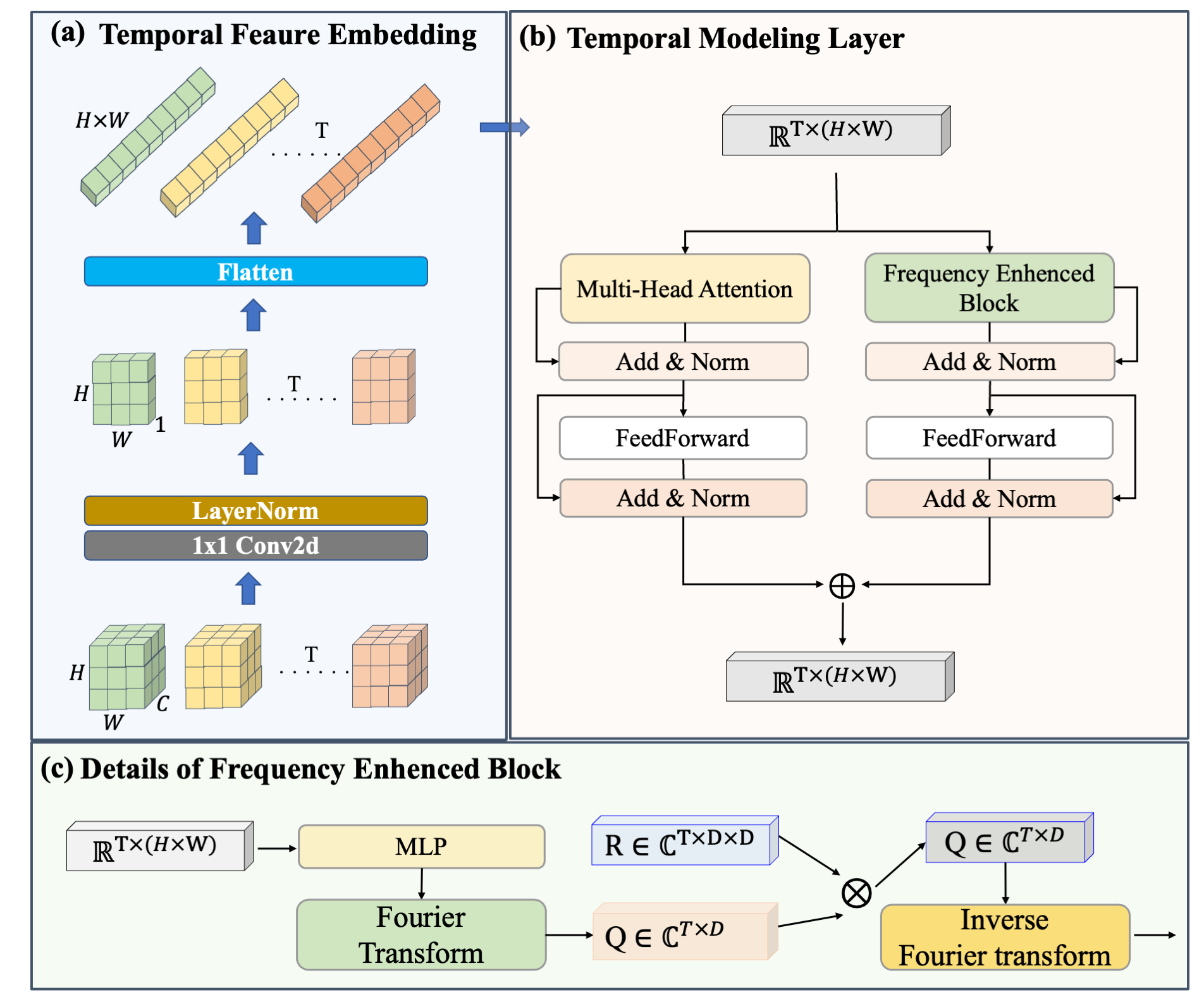}
\caption{Detailed procedure for temporal modeling layer. The spatiotemporal features are first coded through the temporal feature embedding shown in (a), and then entered into the temporal modeling layer in (b) to learn the temporal evolution law of the echo. (c) details the specific structure of the frequency enhanced block.}
\label{fig_ft}
\end{figure}
The temporal evolution of radar echoes can reflect the characteristics of different stages of weather systems. Fine-grained modeling of the temporal evolution pattern aids in gaining a deeper understanding of the intrinsic properties of radar echoes, leading to improved accuracy in radar echo extrapolation.

\begin{figure}
\centering
\includegraphics[width=1.0\linewidth]{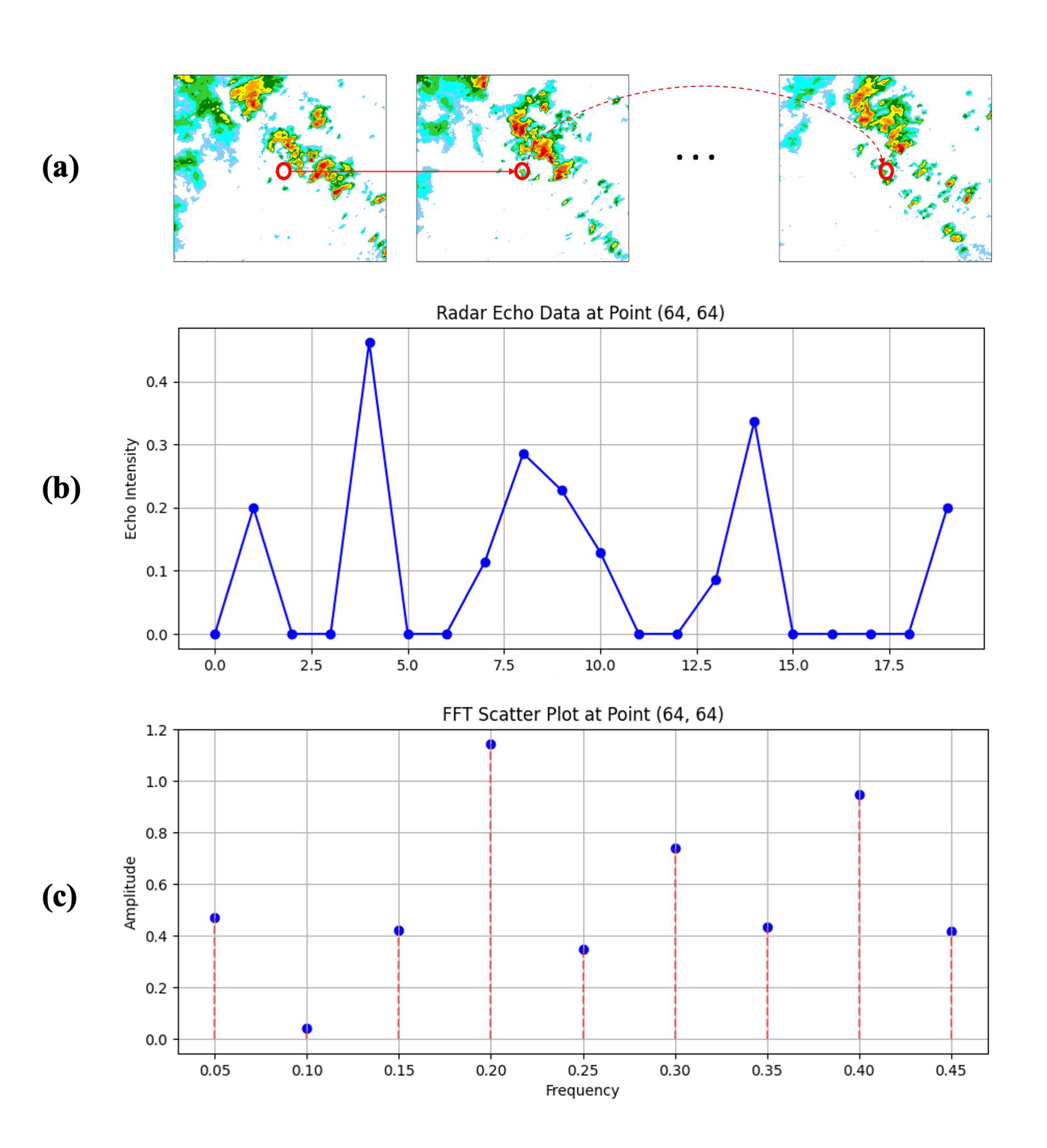}
\caption{(a) A precipitation case in the Hong Kong region, with the selected observation point marked by a red circle. (b) Temporal variation of radar echo intensity at the annotated observation point. (c) Frequency domain representation of radar echo intensity at the annotated observation point.}
\label{fig_ts}
\end{figure}

As widely recognized, time series data can be modeled using both the time domain and the frequency domain. The time domain represents how a signal changes over time, providing insights into the signal's temporal fluctuations. However, certain periodic patterns within signals may be challenging to observe in the time domain, especially when these patterns are very subtle or mixed with other noise or variations. Frequency domain analysis, such as Fourier transform, serves as a valuable tool for converting signals into a representation of their frequency components, which allows us to identify the oscillatory behavior at different frequencies within a signal, even when these patterns are not readily apparent in the time domain. 
In radar echo extrapolation, the radar reflectivity at each individual observation point forms a discrete time series over a period. As illustrated in Fig. \ref{fig_intro}, we are aware that distinct weather patterns undergo a cyclic process, manifested at individual observation points as periodic patterns. However, analyzing this evolutionary pattern solely in the time domain is not sufficient. Frequency domain analysis serves to amplify these periodic components and make them more evident. 
Fig. \ref{fig_ts}(a) illustrates a precipitation case in the Hong Kong region, with the selected observation point marked by a red circle. The temporal variation of radar echo intensity at the annotated observation point is depicted in Fig. \ref{fig_ts}(b), and its frequency domain representation, obtained through Fourier transform, is shown in Fig. \ref{fig_ts}(c). In the frequency domain representation of the signals, the contribution of each frequency component becomes visibly apparent. By performing Fourier transforms on each observation point and analyzing the periodic patterns at each point, complemented by time-domain analysis, we gain valuable insights into the evolving patterns of radar echoes across the entire region.

Therefore, to explore the temporal evolution pattern of radar echoes, it is necessary to supplement the sole time-domain analysis with frequency-domain analysis. First, we map the input of the SFT-Block to the temporal domain through temporal feature embedding(FE(T) in Fig. \ref{fig_overall_architecture}), resulting in a temporal representation denoted as $z_t \in \mathbb{R}^{T \times (H \times W)}$. The detailed process is depicted in Fig. \ref{fig_ft}(a). The original feature $z \in \mathbb{R}^{T \times C \times H \times W}$ is first passed through a 2D convolution with a kernel size of 1 to reduce the number of channels to 1. Then, it undergoes normalization and flattening operations to obtain $z_t \in \mathbb{R}^{T \times (H \times W)}$. This process can be represented by the following formula:
\begin{equation}
     z_t = \operatorname{Flatten}\left(\operatorname{Norm}\left(\operatorname{Conv2d_{1 \times 1}}\left(z\right)\right)\right)
\end{equation}
As shown in Fig. \ref{fig_ft}(b), the temporal feature embedding $z_t$ then undergoes interaction of temporal information through a traditional self-attention to learn the temporal evolution patterns. Additionally, to model the lifecycle of radar echoes, frequency enhanced block is designed to assist temporal multihead self-attention. The formula representation is as follows:
\begin{equation}
    z_{t}^{'}=\mathrm{LN}\left(\mathrm{MSA}\left(z_{t}\right)\right)+z_{t}
\end{equation}
\begin{equation}
    z_{t}^{''}=\mathrm{LN}\left(\mathrm{FEB}\left(z_{t}\right)\right)+z_{t} 
\end{equation}
\begin{equation}
    f_{t} = \mathrm{ChannelCopy}\left(z_{t}^{'}+ z_{t}^{''},C\right)
\end{equation}
where MSA stands for multihead self-attention, and FEB stands for frequency enhanced block. Channel Copy refers to the operation of duplicating the channel dimension of the temporal features C times, resulting in enriched spatiotemporal representation $f_t$. The specific implementation of FEB is illustrated in Fig. \ref{fig_ft}(c). The temporal embedding $z_t$ undergoes a MLP followed by a Fourier transform, resulting in the frequency domain representation Q in complex space. Subsequently, Q is multiplied with a randomly initialized complex matrix R and finally mapped back to the time domain through inverse Fourier transform. The FEB module is defined as follows:
\begin{equation}
    \mathrm{FEB}(z_t) = \mathcal{F}^{-1}(\mathcal{F}(\mathrm{MLP}(z_t)) \odot R)
\end{equation}
where $R \in \mathbb{C}^{(H\times W)\times (H \times W) \times D}$ is a randomly initialized parameterized kernel. For $Y = Q \odot R$, with $Y \in \mathbb{C}^{T \times (H \times W)}$, the operator 
$\odot$ is defined as $Y_{m, d_o}=\sum_{d_i=0}^D Q_{m, d_i} \cdot R_{d_i, d_o, m}$, where $d_i = 1, 2, ..., D$ represents the input channels, and $d_o = 1, 2, ..., D$ represents the output channels.

\subsection{Prediction-Reconstruction Joint Training Paradigm} \label{sec_method:Restruction}
{When meteorologists make weather forecasts, they often analyze the spatiotemporal characteristics of the radar echo sequence based on their knowledge of meteorology, while also reviewing historical observations to make predictions. To prevent the model from excessively forgetting the original information of historical frames while learning spatiotemporal dynamics, we design the reconstrction module and propose a prediction-reconstruction joint training paradigm.} 

The reconstruction module utilizes the even-frame features of ${\boldsymbol{\mathcal{H}'}}$ and the motion pattern ${\boldsymbol{\mathcal{D}'}}$ to recover the odd-frame features in the initial embedding ${\boldsymbol{\mathcal{H}}}$. {The reason we introduce motion information $\boldsymbol{\mathcal{D}}$ and use it together with features from even frames $\boldsymbol{\mathcal{H'}_{even}}$ to reconstruct the original information from odd frames $\boldsymbol{\mathcal{H}_{odd}}$ is that radar echo data exhibits strong temporal dependencies. The evolution of echoes from one frame to the next carries crucial information about weather patterns. Combining features from even frames with motion information allows us to reconstruct features from odd frames by considering temporal consistency between frames. This approach enables the model to better capture the changing dynamics of echoes over time, which is crucial for accurate predictions. Furthermore, radar echo data exhibits strong temporal redundancy\cite{tong2022videomae}, even by reconstructing only odd frames, we can restore the majority of historical information.} 

As shown at the bottom of Fig. \ref{fig_overall_architecture}, the dissimilarity between the odd-frame and even-frame features of the initial embedding ${\mathcal{H}}$ forms the initial motion pattern ${\mathcal{D}}$. Then ${\mathcal{D}}$ undergoes feature extraction through the motion pattern mining module, which consists of N Swin Transformer Blocks, with the first N-1 blocks sharing model weights with the spatial refinement layers of the SFT-Blocks. This design not only reduces the model's complexity and storage requirements but also enables the model to comprehensively consider the spatial distribution and motion evolution of the echoes at the same level, capturing the features of the echo sequences more comprehensively. The last layer of the Swin Transformer Block operates independently without weight sharing and is responsible for extracting spatial and motion features from the echo sequence individually. This design allows the model to perform personalized feature extraction for specific tasks in the last layer, further optimizing the performance and adaptability of the model. The formula is represented as follows:
\begin{equation}
    {\mathcal{D}}^{p}= MPM\left({\mathcal{D}}^{p-1}\right)
\end{equation}
where $M < p \leq M+N$ represents the layer index in the model; MPM is short for motion pattern mining; ${\mathcal{D}}^M$ corresponds to ${\mathcal{D}}$, and the final extracted echo motion pattern is denoted as ${\mathcal{D}}' = {\mathcal{D}}^{M+N}$. 

\begin{figure}
\centering
\includegraphics[width=1.0\linewidth]{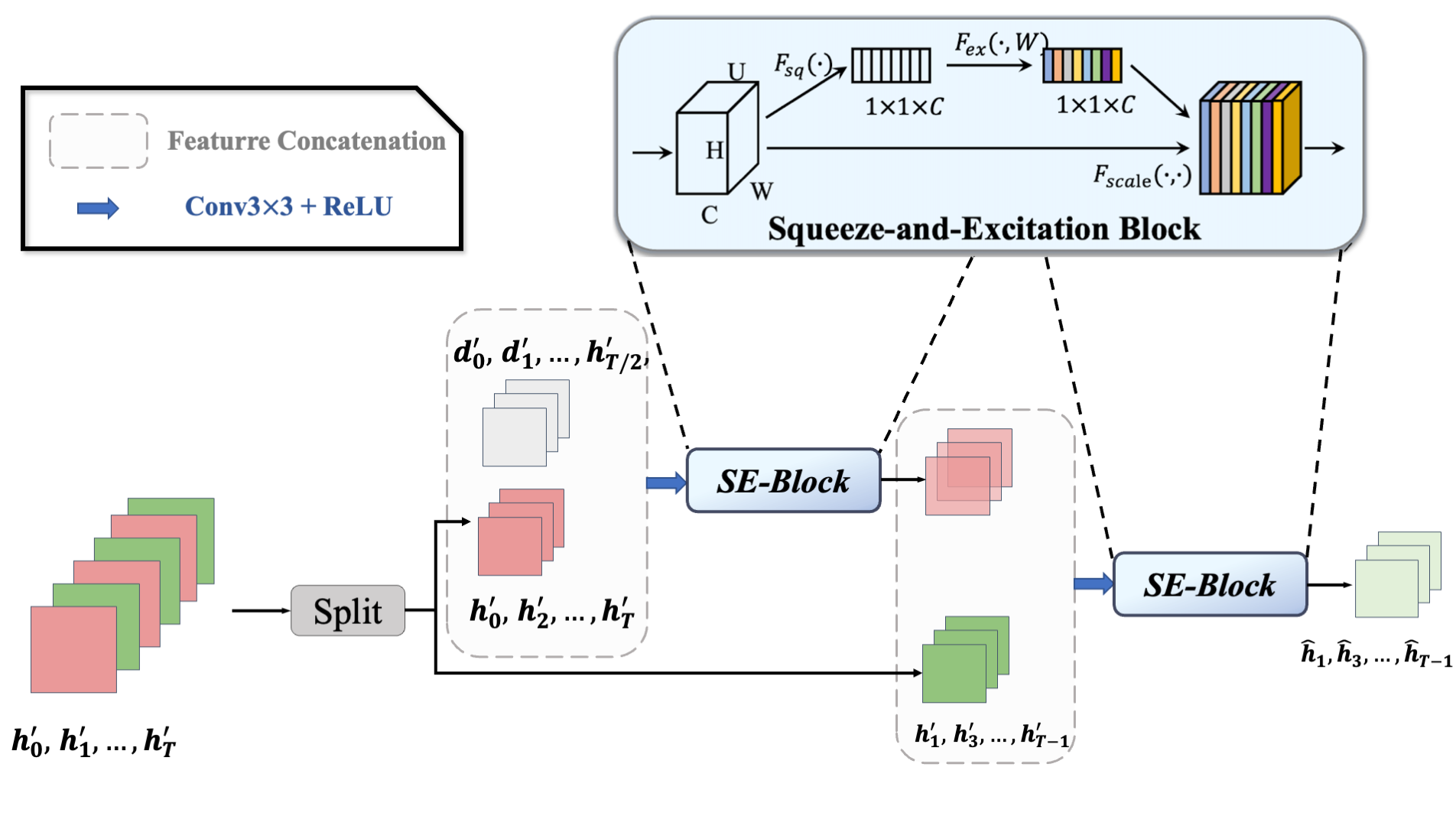}
\caption{The Reconstruction Module. The echo motion pattern features $D'$  contain changes in the spatial morphology between neighboring echo frames in the time dimension; therefore, even frames and motion information are used to recover the features of odd frames to facilitate the learning of echo dynamics.}
\label{fig_res}
\end{figure}

Then, ${\mathcal{D}}'$ and ${\mathcal{H}}'$ enter the reconstruction module, which is illustrated in detail in Fig. \ref{fig_res}.
The even-frame features and echo motion patterns are stacked along the channel dimension. A convolution operation is then applied to reduce the number of channels, followed by the SE-Block to incorporate channel attention. This process yields an initial estimation of the original odd-frame features. Finally, this feature is concatenated with the odd-frame features extracted by the SFT-Block, and another convolution layer is utilized to further reduce the channel dimension. The final odd-frame reconstruction is obtained by applying the channel attention mechanism. The formula representation is as follows:
\begin{equation}
\boldsymbol{{\hat{\mathcal{X'}}}}_{odd}=SE\left(\operatorname{\sigma}\left(\operatorname{Conv2d}\left(\left[\boldsymbol{{\mathcal{H'}}}_{even},{\boldsymbol{\mathcal{D}'}}\right]\right)\right)\right) 
\end{equation}
\begin{equation}
    {\mathcal{H'}}_{even}, {\mathcal{H'}}_{odd} = \{h_{2i}^{'M+N}\},  \{h_{2i+1}^{'M+N}\}
\end{equation}
\begin{equation}
    \boldsymbol{{\hat{\mathcal{X}}}}_{odd}=SE(\operatorname{\sigma}(\operatorname{Conv2d}([\boldsymbol{{\mathcal{H'}}}_{odd},\boldsymbol{{\hat{\mathcal{X'}}}}_{odd}])))
\end{equation}
By utilizing the initial odd-frame as supervision for the reconstruction process, we obtain the reconstruction loss:
\begin{equation}
 \mathcal{L}_{\text{recon}} = \frac{1}{2T} \sum_{t-T+1}^{t-\lfloor \frac{T-1}{2} \rfloor} \| \hat{h}_{2i+1} - h_{2i+1} \|_2 
\end{equation}
where \( h_{2i+1} \) represents the ground truth odd-frame feature and \( \hat{h}_{2i+1} \) represents the reconstructed odd-frame feature at time step \( 2i + 1 \). The sum is taken over the range from \( t-T+1 \) to \( t-\lfloor \frac{T-1}{2} \rfloor \), indicating the frames used for supervision. The \( \| \cdot \|_2 \) represents the L2 norm, which measures the square difference between the ground truth and reconstructed features.

The final prediction result of the forecasting module is denoted as $\hat{Y} = \{\hat{y}\}_t^{t+T^{'}}$. To quantify the disparity between the predicted values and the ground truth, we utilize the Mean Squared Error (MSE) loss, which can be expressed as follows:
\begin{equation}
    \boldsymbol{\mathcal{L}}_{f} = \frac{1}{T^{'}} \sum_{t+1}^{t+T^{'}}\left(\hat{y}^{i} - y^{i}\right)^{2}
\end{equation}

During the training process, the model is optimized by utilizing a joint loss, which is expressed as follows:
\begin{equation}
 \mathcal{L} =\mathcal{L}_{\text{f}} + \lambda \mathcal{L}_{\text{recon}} 
\end{equation}
where \( \lambda \) is the weight controlling the two losses. \( \mathcal{L}_{\text{recon}}  \) represents the reconstruction loss, as mentioned earlier, and $\mathcal{L}_{\text{f}}$ denotes the forecasting loss, which measures the discrepancy between the predicted radar echo intensity (\( \hat{{\mathcal{Y}}} \)) and the ground truth (\( {\mathcal{Y}} \)). The joint loss combines these two components, allowing the model to simultaneously optimize for both reconstruction and prediction tasks.

\section{Experiments} \label{sec:experiment}
This section covers the experimental setup and result analysis. We begin with the dataset description in Section \ref{sec:data}, implementation details in Section \ref{subsec:imp}, and evaluation metrics in Section \ref{subsec:eval}. In Section \ref{compare}, we present comprehensive experiments comparing SFTformer with classical echo extrapolation and spatiotemporal prediction methods, and provide visualization results for specific cases. Section \ref{sec:vis} showcases the visualization results of different methods and specific radar echo movements. Section \ref{ablation} conducts ablation experiments to analyze each module's role.

\subsection{Dataset Description} \label{sec:data}
\subsubsection{HKO-7 Dataset}
We evaluate the performance of SFTformer through experiments conducted on the HKO-7 dataset\cite{shi2017deep}, which consists of radar echo data from the Hong Kong Observatory between 2009 to 2015. The radar CAPPI reflectivity images encompass a 512-km$^2$ region centered on Hong Kong, captured from an altitude of 2km above sea level. The images have a resolution of 480$\times$480 and are recorded continuously at 6-minute intervals. The dataset is split into three sets: training, validation, and test, with 812, 50, and 131 days of rainy data, respectively. We have structured our dataset based on this division. To validate the model's performance in predicting time series of different lengths, We sample the raw data using sliding windows of widths 20, 30, and 40 frames(input 10 frames, extrapolat 10 frames, 20 frames and 30 frames, respectively), with a stride of 5 frames. With the window size of 20 frames, we obtain a total of 37,444, 2,211, and 6,109 echo sequences for training, validation, and testing, respectively. For the window size of 30 frames, we obtain a total of 36,785, 2,121, and 5,997 echo sequences for training, validation, and testing, respectively. For the window size of 40 frames, we obtain a total of 36,129, 2,031, and 5,885 echo sequences for training, validation, and testing, respectively.
\subsubsection{ChinaNorth-2021 Dataset}
We also conduct experiments using the ChinaNorth-2021 dataset, which consists of radar reflectivity images captured nationwide from June to August 2021, during the rainy season. The geographical coverage spans from 109.9 to 123.0 degrees East longitude and 32.98 to 46.14 degrees North latitude. We preprocess the raw image by dividing it into a grid of data based on latitude and longitude, which greatly simplifies subsequent operations. Considering the extensive coverage of the national scale, we select a subset of the data with a size of 128$\times$128, specifically focusing on the northern region of China. Additionally, we filtered the samples by selecting those with radar echo intensity exceeding a certain threshold. As a result, we obtained the ChinaNorth-2021 dataset for further analysis and experiments, consisting of 1200 sequences for training, 231 sequences for validation and 159 sequences for testing. Each sequence in the dataset has a length of 20 (10 for input and 10 for extrapolation). 
\subsection{Implementing Details} \label{subsec:imp}
 We use historical observation sequence with a length of 10 to extrapolate future radar echo frames. To evaluate the model's proficiency in short, medium, and long-term precipitation forecasting, for HKO-7, the model is trained to forecast the next 10 frames (1 hour), 20 frames (2 hours), and 30 frames (3 hours) ahead. ChinaNorth-2021 consists solely of radar echo sequences with a length of 20 frames, which enables the model to predict radar echo maps for the next 10 frames (1 hour) exclusively on this dataset. The radar echo images are resized to 128$\times$128 pixels. The Feature Embedding and Forecasting module are configured with a layer count of M set to 4, while the SFT-Block module has a layer count of N set to 8. {During the training phase, we set the weight $\lambda$ for $\mathcal{L}_{\text{recon}}$ to 0.01.} {In the inference phase, size of the input sequence to the model is $10 \times 1 \times 128 \times 128$; after Feature Embedding, the feature size becomes $10 \times 64 \times 32 \times 32$; following SFT-Blocks, the feature size remains unchanged, and ultimately, after upsampling in the Forecasting module, the feature is restored to $10 \times 1 \times 128 \times 128$.}

The proposed model is trained using a single NVIDIA GeForce RTX 3090 and an Intel(R) Xeon(R) Gold 6226R CPU, with PyTorch as the training framework. During the training phase, the model is trained using the Adam optimizer with betas $(\beta_1, \beta_2) = (0.9, 0.999)$ and the OneCycleLR learning rate scheduler. The maximum learning rate is set to 0.0001. 

\subsection{Evaluation Metrics} \label{subsec:eval}
For radar echo extrapolation, the evaluation of predicted results involves metrics such as Critical Success Index (CSI), Gilbert Skill Score (GSS), and Heidke Skill Score (HSS). These metrics serve to assess the consistency and accuracy between the predicted and observed results, offering a quantitative evaluation of the model's performance. For our analysis, we choose several rain-rate thresholds (0.5mm/h, 2mm/h, 5mm/h, 10mm/h, 30mm/h) to calculate CSI, GSS, and HSS, assessing the model's predictive capability for different intensities of precipitation.

To calculate these metrics, we begin by transforming the predicted pixel values and ground truth into binary values (0 or 1) using the threshold $\tau$, which enables us to determine the true positive (TP), false positive (FP), false negative (FN), and true negative (TN) counts. Based on these counts, the metrics are calculated as follows:\\
\textbf{Critical Success Index (CSI)}: The Critical Success Index (CSI) is a widely used metric that quantifies the proportion of correctly predicted events among the total events, which is calculated as:
\begin{equation}
   CSI = \frac{TP}{TP + FN + FP}
\end{equation}
\par
The CSI value varies between 0 and 1, with values approaching 1 indicating a higher level of agreement between the predicted and observed results.\\
\textbf{Gilbert Skill Score (GSS)}: GSS is a metric used to assess the accuracy of predictions for extreme events, often employed in precipitation event forecasting. The GSS is calculated as:\par
\begin{equation}
   GSS = \frac{(TP - E_{TP}) + (TN - E_{TN})}{TP + FN + FP + TN - E_{TP} - E_{TN}}
\end{equation}

where $E_{TP}$ and $E_{TN}$ represent the expected counts of TP and TN, respectively, which are typically assumed based on relative frequencies. The GSS value spans from -1 to 1, with values nearing 1 signifying higher accuracy in predicting extreme events.\\
\textbf{Heidke Skill Score (HSS)}: HSS is a widely used metric for evaluating the skill of predictions relative to random predictions. It is calculated as:\par
{\small{
\begin{equation}
\begin{aligned}
   &HSS = \\
   &\frac{2 \times (TP \times TN - FN \times FP)}{(TP + FN)(FN + TN) + (TP + FP)(FP + TN)}
\end{aligned}
\end{equation}
}}
\par
The HSS value varies between 0 and 1, with higher values indicating better skill in predictions compared to random predictions. 

These evaluation metrics provide objective measures of the model's predictive ability and accuracy, offering insights into the consistency between predicted and observed results.
\subsection{Comparions with State-of-the-Art Forecasting Methods} \label{compare}
To demonstrate the superiority of SFTformer, we compare it with several classical spatiotemporal prediction models(SimVP\cite{gao2022simvp},MAU\cite{chang2021mau},PredRNN\cite{wang2017predrnn}, PredRNN++\cite{wang2018predrnn++}) and deep learning based radar echo extrapolation models(ConvLSTM\cite{shi2015convolutional},TrajGRU\cite{shi2017deep}, REMNet\cite{jing2022remnet}). All models undergo training and testing using the identical optimizer and learning rate decay strategy as the SFTformer, as described in Section \ref{subsec:imp}.

 TABLE \ref{tab:contrast} and TABLE \ref{tab:contrast2} present the quantitative evaluation results of different methods on HKO-7 and ChinaNorth-2021, respectively. We comprehensively compare SFTformer with other advanced forecasting methods. Using the CSI and HSS metrics on the HKO-7 dataset, we assess the model's performance at 1-hour, 2-hour, and 3-hour prediction durations and different radar echo reflectivity thresholds. On the ChinaNorth-2021 dataset, we evaluate the model's performance at a 1-hour prediction duration and a radar echo reflectivity threshold of 0.5mm/h using CSI, HSS, and GSS metrics. For various prediction durations and reflectivity thresholds, SFTformer excels in comparative experiments. In the 1-hour prediction duration, compared to the generic method SimVP\cite{gao2022simvp}, SFTformer achieves higher CSI and HSS values at all reflectivity thresholds, reaching 0.625 and 0.733 on the 0.5mm/h threhold, respectively, significantly outperforming other comparison methods. This pattern is consistent across both the HKO-7 and ChinaNorth-2021 datasets, demonstrating the model's outstanding performance in short-term predictions. As the prediction duration extends to 2 hours and 3 hours on the HKO-7 dataset, the proposed method continues to maintain a leading position in CSI and HSS metrics, obtaining the highest values at all reflectivity thresholds. In contrast, other generic and precipitation forecasting methods show varying performance at different reflectivity thresholds and exhibit some performance gaps compared to SFTformer, which highlights the robustness and superior performance of our method in long-term predictions. 
In summary, SFTformer demonstrates outstanding performance across different prediction durations and reflectivity levels, proving its effectiveness in radar echo extrapolation.

\begin{table*}[htbp]
  \centering
  \caption{Comparions with State-of-the-Art Forecasting Methods on HKO-7. Bolded indicates the best, underlined the second best. r represents radar echo reflectivity. "Generic" refers to general spatiotemporal prediction methods, while "Precipitation" pertains to precipitation forecasting methods.}
  \resizebox{\linewidth}{!}{
    \begin{tabular}{c|c|l|ccccc|ccccc}
    \toprule
    \multicolumn{1}{c|}{\multirow{2}[4]{*}{Duration}} & \multicolumn{2}{c|}{\multirow{2}[4]{*}{Methods}} & \multicolumn{5}{c|}{CSI}              & \multicolumn{5}{c}{HSS} \\
\cmidrule{4-13}          & \multicolumn{2}{c|}{} & \multicolumn{1}{c|}{r $\geq$ 0.5} & \multicolumn{1}{c|}{r $\geq$ 2} & \multicolumn{1}{c|}{r $\geq$ 5} & \multicolumn{1}{c|}{r$\geq$ 10} & r $\geq$ 30  & \multicolumn{1}{c|}{r$\geq$ 0.5} & \multicolumn{1}{c|}{r$\geq$ 2} & \multicolumn{1}{c|}{r $\geq$ 5} & \multicolumn{1}{c|}{r$\geq$ 10} & r$\geq$ 30 \\
    \midrule
    \multirow{8}[6]{*}{1h} & \multirow{4}[2]{*}{Generic} & SimVP\cite{gao2022simvp} & 0.605 & 0.49  & 0.360  & 0.246 & 0.128 & 0.719 & \underline{0.626} & 0.497 & 0.363 & 0.200 \\
          &       & MAU\cite{chang2021mau}   & 0.569 & \underline{0.495} & 0.326 & 0.209 & 0.096 & 0.688 & 0.591 & 0.457 & 0.313 & 0.154 \\
          &       & PredRNN\cite{wang2017predrnn} & 0.606 & 0.482 & 0.338 & 0.211 & 0.090  & 0.721 & 0.619 & 0.472 & 0.317 & 0.145 \\
          &       & PredRNN++\cite{wang2018predrnn++} & \underline{0.609} & 0.480  & 0.342 & 0.221 & 0.102 & \underline{0.723} & 0.618 & 0.477 & 0.331 & 0.165 \\
\cmidrule{2-13}          & \multicolumn{1}{c|}{\multirow{4}[4]{*}{Precipitation}} & ConvLSTM\cite{shi2015convolutional} & 0.600   & 0.461 & 0.311 & 0.188 & 0.077 & 0.716 & 0.599 & 0.439 & 0.284 & 0.124 \\
          &       & TrajGRU\cite{shi2017deep} & 0.543 & 0.461 & \underline{0.371} & \underline{0.287} & \underline{0.186} & 0.661 & 0.461 & 0.371 & 0.287 & 0.186 \\
          &       & Remnet\cite{jing2022remnet} & 0.575 & 0.481 & 0.370  & 0.281 & 0.162 & 0.690  & 0.616 & \underline{0.510}  & \underline{0.413} & \underline{0.260} \\
          &       & \textbf{\textit{Ours}}  & \textbf{0.625} & \textbf{0.503} & \textbf{0.377} & \textbf{0.288} & \textbf{0.209} & \textbf{0.733} & \textbf{0.636} & \textbf{0.518} & \textbf{0.422} & \textbf{0.328} \\
    \midrule
    \multirow{8}[6]{*}{2h} & \multirow{4}[2]{*}{Generic} & SimVP\cite{gao2022simvp} & 0.492 & 0.339 & 0.196 & 0.100   & 0.035 & 0.611 & 0.459 & 0.285 & 0.153 & 0.057 \\
          &       & MAU\cite{chang2021mau}   & 0.452 & 0.310  & 0.188 & 0.100   & 0.035 & 0.571 & 0.425 & 0.275 & 0.155 & 0.058 \\
          &       & PredRNN\cite{wang2017predrnn} & 0.489 & 0.338 & 0.199 & 0.103 & 0.04  & 0.608 & 0.459 & 0.292 & 0.159 & 0.064 \\
          &       & PredRNN++\cite{wang2018predrnn++} & \underline{0.494} & 0.339 & 0.198 & 0.100   & 0.034 & \underline{0.612} & 0.46  & 0.29  & 0.154 & 0.056 \\
\cmidrule{2-13}          & \multicolumn{1}{c|}{\multirow{4}[4]{*}{Precipitation}} & ConvLSTM\cite{shi2015convolutional} & 0.476 & 0.320  & 0.197 & 0.104 & 0.028 & 0.596 & 0.438 & 0.292 & 0.164 & 0.047 \\
          &       & TrajGRU\cite{shi2017deep} & 0.443 & 0.363 & \textbf{0.278} & \textbf{0.203} & \underline{0.105}
 & 0.551 & \underline{0.488} & \textbf{0.401} & \textbf{0.311} & \underline{0.171} \\
          &       & Remnet\cite{jing2022remnet} & 0.473 & \underline{0.366} & \underline{0.269}& 0.181 & 0.093 & 0.582 & \underline{0.488} & 0.373 & 0.276 & 0.154 \\
          &       & \textbf{\textit{Ours}}  & \textbf{0.519} & \textbf{0.395} & \textbf{0.278} & \underline{0.198} & \textbf{0.123} & \textbf{0.626} & \textbf{0.519} & \underline{0.397} & \underline{0.302} & \textbf{0.199} \\
    \midrule
    \multirow{8}[6]{*}{3h} & \multirow{4}[2]{*}{Generic} & SimVP\cite{gao2022simvp} & \underline{0.413} & 0.244 & 0.121 & 0.053 & 0.017 & \underline{0.527} & 0.341 & 0.181 & 0.082 & 0.029 \\
          &       & MAU\cite{chang2021mau}   & 0.351 & 0.194 & 0.099 & 0.042 & 0.008 & 0.459 & 0.275 & 0.149 & 0.067 & 0.014 \\
          &       & PredRNN\cite{wang2017predrnn} & 0.387 & 0.232 & 0.121 & 0.053 & 0.012 & 0.498 & 0.323 & 0.181 & 0.083 & 0.021 \\
          &       & PredRNN++\cite{wang2018predrnn++} & 0.392 & 0.241 & 0.123 & 0.053 & 0.009 & 0.504 & 0.338 & 0.184 & 0.085 & 0.016 \\
\cmidrule{2-13}          & \multicolumn{1}{c|}{\multirow{4}[4]{*}{Precipitation}} & ConvLSTM\cite{shi2015convolutional} & 0.335 & 0.173 & 0.085 & 0.039 & 0.01  & 0.437 & 0.243 & 0.127 & 0.061 & 0.016 \\
          &       & TrajGRU\cite{shi2017deep} & 0.386 & 0.308 & \textbf{0.227} & \textbf{0.159} & \underline{0.074} & 0.483 & \underline{0.418} & \textbf{0.331} & \textbf{0.248} & \textbf{0.121} \\
          &       & Remnet\cite{jing2022remnet} & 0.412 & \underline{0.310}  & 0.207 & 0.132 & 0.06  & 0.512 & \underline{0.418} & 0.299 & 0.203 & 0.098 \\
          &       & \textbf{\textit{Ours}}  & \textbf{0.449} & \textbf{0.331} & \underline{0.223} & \underline{0.150}  & \textbf{0.073} & \textbf{0.546} & \textbf{0.441} & \underline{0.323} & \underline{0.231} & \underline{0.120} \\
    \bottomrule
    \end{tabular}%
    }
  \label{tab:contrast}%
\end{table*}%

% Table generated by Excel2LaTeX from sheet 'Contrast20 huabei (2)'

\begin{table}[htbp]
  \centering
  \caption{Comparions with State-of-the-Art Forecasting Methods on ChinaNorth-2021 dataset. "Generic" refers to general spatiotemporal prediction methods, while "Precipitation" pertains to precipitation forecasting methods. Bolded indicates the best, underlined the second best. }
    \begin{tabular}{c|c|c|c|c}
    \toprule
    \multicolumn{2}{c|}{Methods} & CSI   & GSS & HSS \\
    \midrule
    \multirow{4}[1]{*}{Generic} & SimVP & 0.362 & 0.352 & 0.512 \\
          & MAU   & \underline{0.391} & 0.379 & 0.542 \\
          & PredRNN & 0.369 & 0.357 & 0.520 \\
          & PredRNN++ & \underline{0.391} & \underline{0.380}  & \underline{0.543} \\
    \midrule
    \multirow{4}[2]{*}{Precipitation} & ConvLSTM & 0.356 & 0.345 & 0.506 \\
          & Remnet & 0.318 & 0.306 & 0.465 \\
          & TrajGRU & 0.341 & 0.333 & 0.486 \\
          & \textit{\textbf{Ours}} & \textbf{0.398} & \textbf{0.387} & \textbf{0.552} \\
    \bottomrule
    \end{tabular}%
  \label{tab:contrast2}%
\end{table}%

\subsection{Results Visualization}
\label{sec:vis}
\begin{figure}
\centering
\includegraphics[width=1.0\linewidth]{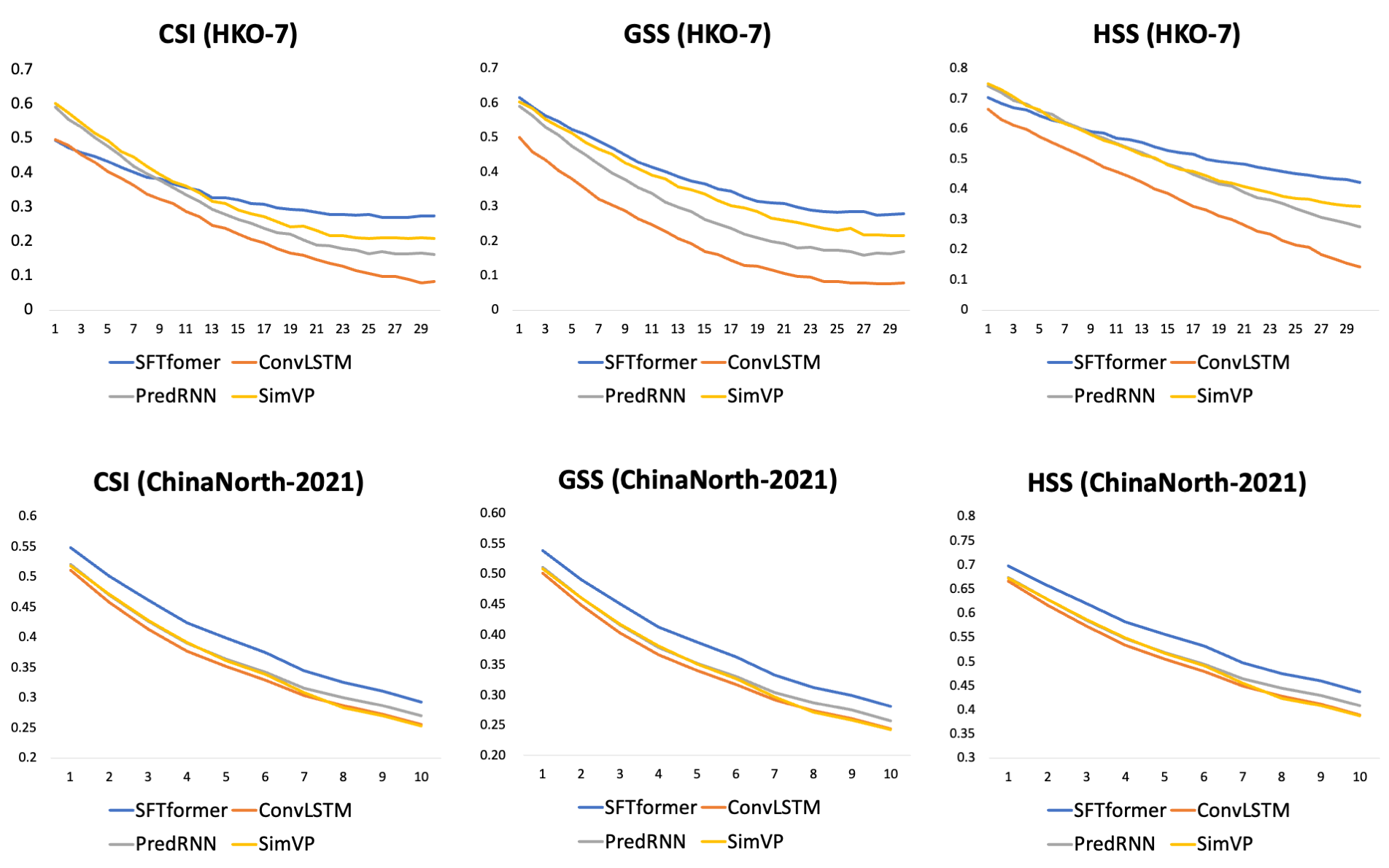}
\caption{CSI, GSS, HSS changing curves along different extrapolation timesteps of 5 extrapolation methods on HKO-7 and ChinaNorth-2021.}
\label{fig_csi}
\end{figure}
On both datasets, we compare the proposed method (SFTformer) with several other methods (ConvLSTM\cite{shi2015convolutional}, PredRNN\cite{wang2017predrnn}, SimVP\cite{gao2022simvp}) in terms of the CSI, HSS, and GSS metrics across different extrapolation time steps. Fig. \ref{fig_csi} displays the relative performance of these methods on the extrapolation time steps, (a) for the HKO-7 dataset(3h forecasting) and (b) for ChinaNorth-2021(1h forecasting). As the extrapolation time steps increase, all extrapolation methods exhibit a long-term decay trend, but our method (SFTformer) demonstrates a slower decay compared to the others.

On the three metrics of the HKO-7 dataset, ConvLSTM\cite{shi2015convolutional} exhibits a promising initial performance, but as the extrapolation time steps increase, the CSI value gradually decreases and decays fast after the 13th time step. SimVP\cite{gao2022simvp} shows a similar trend, albeit with a slower decay after the 13th time step compared to ConvLSTM\cite{shi2015convolutional}. PredRNN\cite{wang2017predrnn} make structural improvements to mitigate long-term decay, maintaining relatively stable performance across the entire range of extrapolation time steps. Overall, although our method (SFTformer) experiences some decay during long-term extrapolation, it exhibits a slower decay compared to the other methods, indicating a relatively stable predictive accuracy.\par
\begin{figure*}
\centering
\includegraphics[width=\linewidth]{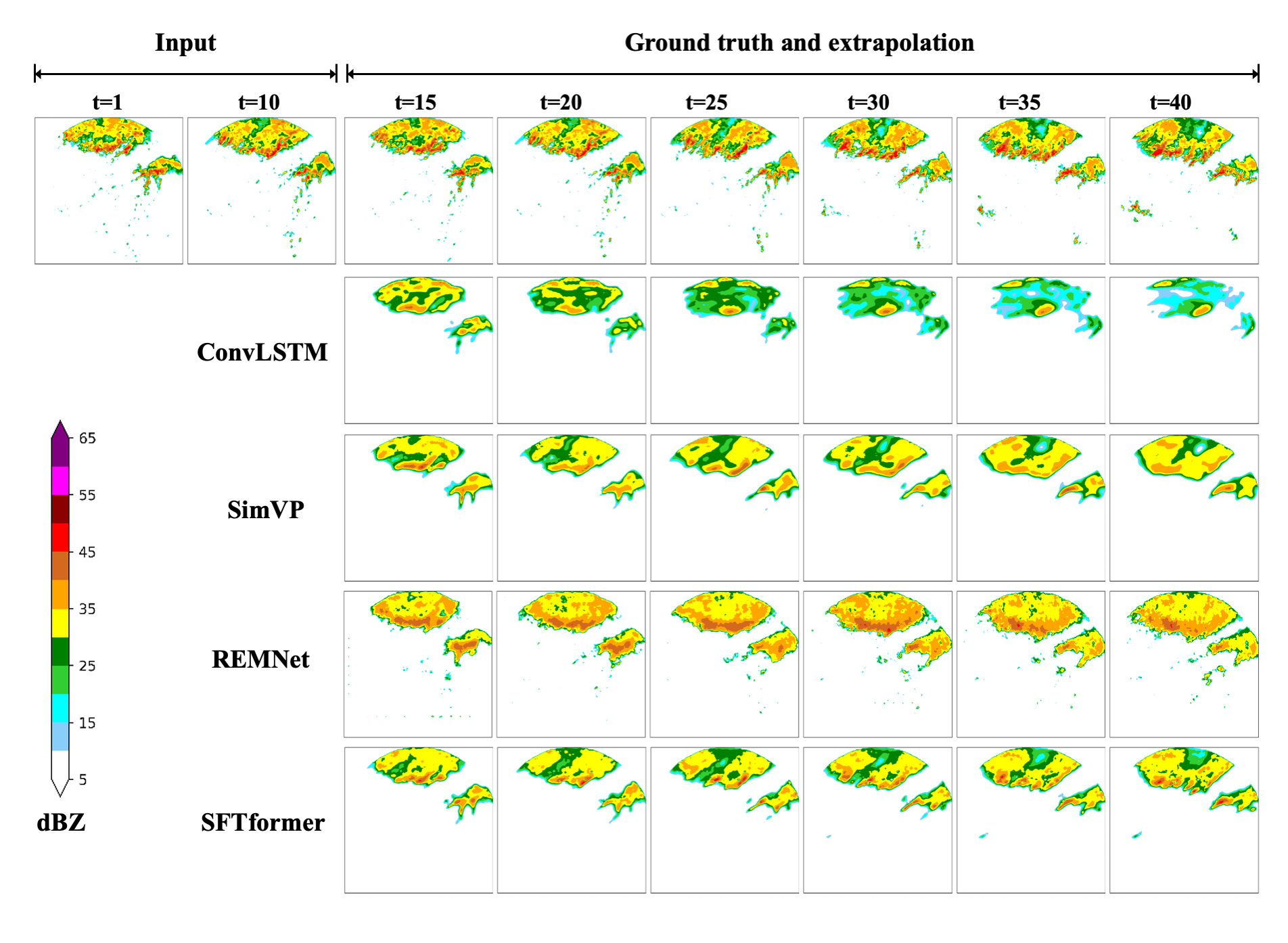}
\caption{Comparision of different models on extrapolation results of a precipitation event that happened in Hong Kong. The models perform 3h forecasting observing 1h historical echo frames on the Hong Kong samples.}
\label{fig_compare}
\end{figure*}

\begin{figure*}
\centering
\includegraphics[width=0.9\linewidth]{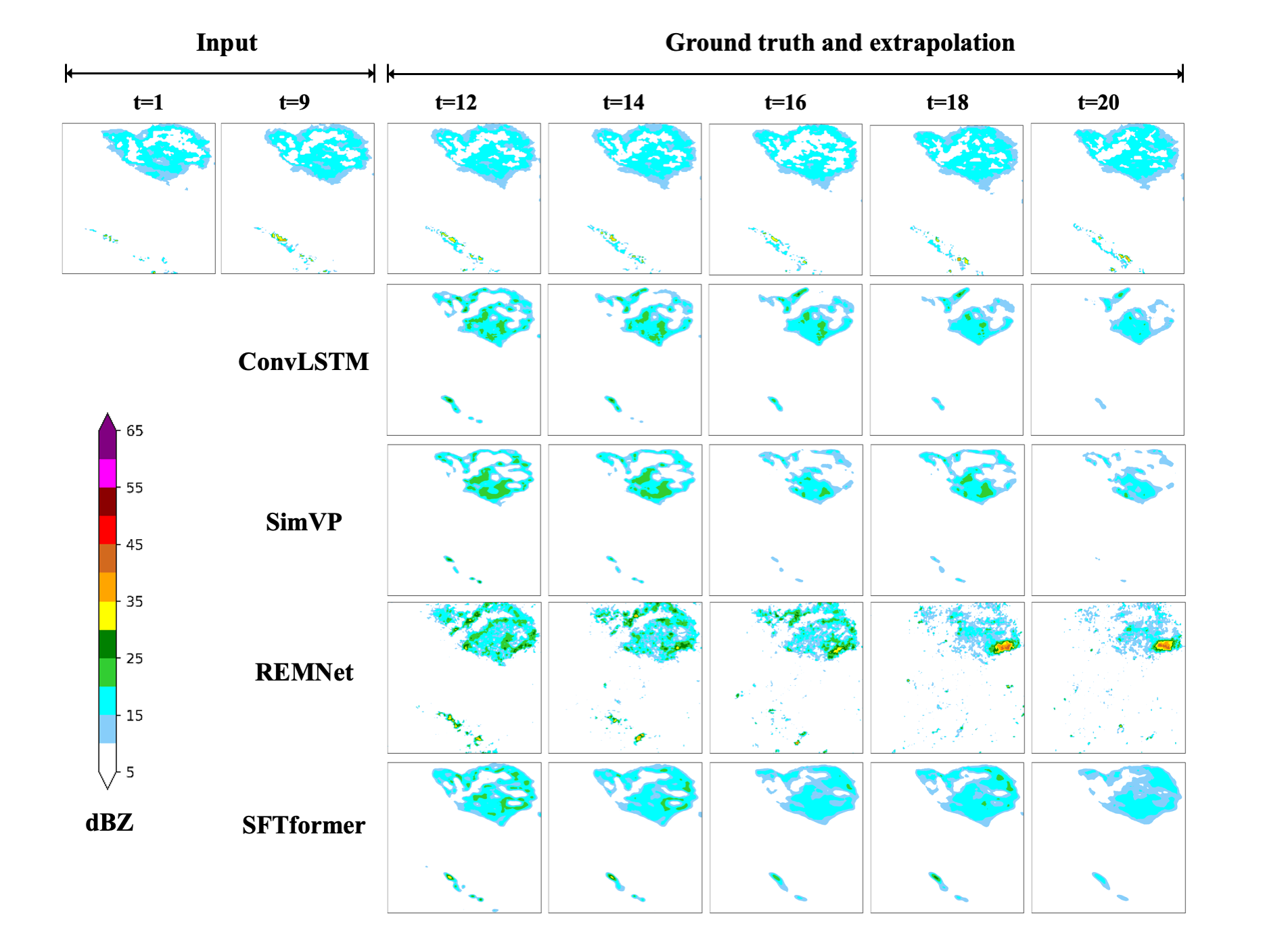}
\caption{Comparision of different models on extrapolation results of a precipitation event that happened in North of China. The models perform 1h forecasting observing 1h historical echo frames on the ChinaNorth samples.}
\label{fig_compare2}
\end{figure*}

{Fig. \ref{fig_compare} and Fig. \ref{fig_compare2}} present a comparison of extrapolation results of different methods on two samples, one from HKO-7 and the other from ChinaNorth-2021. Fig. \ref{fig_compare} displays a convective weather event characterized by both concentrated and scattered distribution, with relatively rapid evolution. The first row shows the historical observed sequence and the ground truth for prediction. Analyzing the first three columns of the extrapolation results in Fig. \ref{fig_compare}, it becomes apparent that within the initial hour, our method's extrapolation results do not demonstrate a significant advantage over the other methods. ConvLSTM\cite{shi2015convolutional}, for instance, performs reasonably well, albeit primarily in short-term predictions. Nonetheless, as we extend the predictions to longer time frames, they encounter challenges in capturing long-term temporal evolution patterns and spatial morphology. Consequently, we observe a gradual blurring in their predictions. SimVP\cite{gao2022simvp}'s prediction results are slightly better than ConvLSTM, but it still loses more details in long-term forecasts. In long-term forecasting, our model better maintains the shape corresponding to the ground truth. In Fig. \ref{fig_compare2}, the radar echo sequence exhibits a heart shape, with slight variations occurring only within the interior of the heart shape over time. This example aims to validate the models' abilities to capture the fine details of the echo patterns. Up to the 14th time step, all models perform similarly, with our model providing more complete predictions of the internal details of the echo compared to other models. However, after the 14th time step, other models begin to exhibit details loss, particularly evident in the last time step, where ConvLSTM\cite{shi2015convolutional}'s results almost lose the heart shape's form, and the rainfall band in the bottom left corner of the image is barely predicted. Other models also show similar issues to some extent. {However, our method exhibits the issue of over-smoothing in the predicted radar echo images, which is attributed to spatial downsampling and the use of L2 loss. This remains a challenge to be addressed in future research.}

\begin{figure*}
\centering
\includegraphics[width=1.0\linewidth]{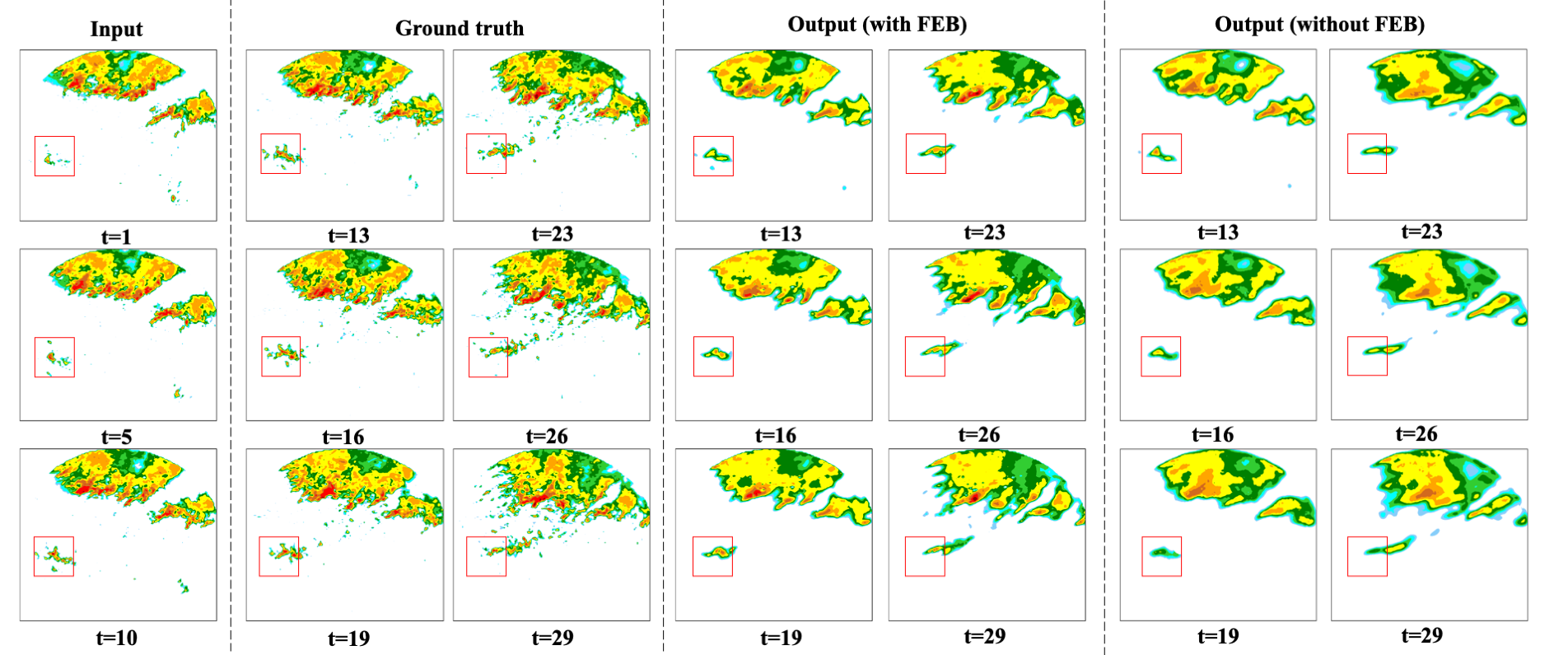}
\caption{The visualization results of a precipitation case with and without the Frequency Enhanced Module (FEM). The region of interest is highlighted by the red rectangular box.}
\label{fig_cyc}
\end{figure*}

Moreover, to visually demonstrate the effectiveness of our proposed method in modeling the temporal evolution patterns and the lifecycle of radar echoes, we have visualized relevant precipitation case, as shown in Fig. \ref{fig_cyc}, where the future echo sequence for the next 2 hours is extrapolated using the historical echo sequence from the preceding hour. The rectangular box indicates the echo cell of interest. In the region highlighted by the rectangular box in the input sequence, we observe a trend of increasing echo intensity and expanding area. This trend persists until the 16th time step in the ground truth. Over time, the echo intensity within the region of interest noticeably decreases, reaching a more significant attenuation by the 29th time step. The output results of the proposed complete model with frequency enhanced module (Output with FEM) effectively capture this trend. From the 13th to the 16th and 19th time steps, there is an enhancement in echo intensity, followed by a gradual decline. This illustrates that our model can adeptly model the temporal evolution patterns of echoes. When frequency domain analysis is not applied, the output results of the model (Output without FEM) show a slight reduction in echo intensity from the 13th to the 16th time step, with a continuous decrease, leading to a slight decline in predictive performance. This underscores the necessity and effectiveness of frequency domain analysis and demonstrates the the model's ability to capture the periodic evolution of each point contributes to better predictions of regional changes.

\subsection{Ablation Studies} \label{ablation}

\begin{table*}[htbp]
  \centering
  \caption{Statistics on the results of the ablation experiment on HKO-7, 3 hours forecasting. r(lowercase) represents radar echo reflectivity. T stands for Temporal Modeling Layer, S stands for Spatial Refinement Layer, ST stands for Spatiotemporal Correlation Layer, and R(uppercase) stands for Reconstruction module.}
  \resizebox{\linewidth}{!}{
    \begin{tabular}{cccc|ccccc|ccccc|ccccc}
\cmidrule{1-19}    \multicolumn{4}{c|}{Modules}  & \multicolumn{5}{c|}{CSI}              & \multicolumn{5}{c|}{GSS}               & \multicolumn{5}{c}{HSS} \\
\cmidrule{1-19}
    \multicolumn{1}{c|}{T} & \multicolumn{1}{c|}{S} & \multicolumn{1}{c|}{ST} & R & \multicolumn{1}{c|}{r$\ge$0.5} & \multicolumn{1}{c|}{r$\ge$2} & \multicolumn{1}{c|}{r$\ge$5} & \multicolumn{1}{c|}{r$\ge$10} & r$\ge$30  & \multicolumn{1}{c|}{r$\ge$0.5} & \multicolumn{1}{c|}{r$\ge$2} & \multicolumn{1}{c|}{r$\ge$5} & \multicolumn{1}{c|}{r$\ge$10} & r$\ge$30  & \multicolumn{1}{c|}{r$\ge$0.5} & \multicolumn{1}{c|}{r$\ge$2} & \multicolumn{1}{c|}{r$\ge$5} & \multicolumn{1}{c|}{r$\ge$10} & r$\ge$30 \\
    \midrule
          &       &       &       & 0.417 & 0.306 & 0.197 & 0.116 & 0.021 & 0.353 & 0.265 & 0.173 & 0.116 & 0.02  & 0.51  & 0.411 & 0.288 & 0.183 & 0.036 \\
          & $\checkmark$     & $\checkmark$     & $\checkmark$     & 0.425 & 0.319 & 0.221 & 0.149 & 0.06  & 0.368 & 0.283 & 0.198 & 0.136 & 0.059 & 0.525 & 0.429 & 0.32  & 0.23  & 0.1 \\
    $\checkmark$     &       & $\checkmark$     & $\checkmark$     & 0.435 & 0.324 & 0.223 & 0.151 & 0.052 & 0.374 & 0.286 & 0.199 & 0.137 & 0.051 & 0.53  & 0.433 & 0.323 & 0.233 & 0.086 \\
    $\checkmark$     & $\checkmark$     &       & $\checkmark$     & 0.435 & 0.323 & 0.22  & 0.146 & 0.049 & 0.372 & 0.285 & 0.197 & 0.133 & 0.047 & 0.528 & 0.433 & 0.32  & 0.226 & 0.081 \\
    $\checkmark$     & $\checkmark$     & $\checkmark$     &       & 0.436 & 0.324 & 0.221 & 0.148 & 0.051 & 0.374 & 0.286 & 0.198 & 0.135 & 0.049 & 0.531 & 0.434 & 0.321 & 0.229 & 0.084 \\
    \midrule
    $\checkmark$     & $\checkmark$     & $\checkmark$     & $\checkmark$     & 0.449 & 0.331 & 0.223 & 0.15  & 0.073 & 0.389 & 0.292 & 0.2   & 0.137 & 0.07  & 0.546 & 0.441 & 0.323 & 0.231 & 0.12 \\
    \bottomrule
    \end{tabular}
    }
  \label{tab:abilation}%
\end{table*}%

\begin{table}[htbp]
  \centering
  \caption{Statistics on the results of the ablation experiment on ChinaNorth-2021. m indicates the mean of the predicted frames, and l indicates the result of the last frame. T stands for Temporal Modeling Layer, S stands for Spatial Refinement Layer, ST stands for Spatiotemporal Correlation Layer, and R stands for Reconstruction module.}
  \resizebox{\linewidth}{!}{
    \begin{tabular}{cccc|cc|cc|cc}
\cmidrule{1-10}    \multicolumn{4}{c|}{Modules}  & \multicolumn{2}{c|}{CSI} & \multicolumn{2}{c|}{GSS} & \multicolumn{2}{c}{HSS} \\
    \midrule
    \multicolumn{1}{c|}{T} & \multicolumn{1}{c|}{S} & \multicolumn{1}{c|}{ST} & R & M     & L     & M     & L     & M     & L \\
    \midrule
          &       &       &       & 0.372 & 0.263 & 0.361 & 0.252 & 0.523 & 0.400 \\
          & $\checkmark$     & $\checkmark$     & $\checkmark$     & 0.381 & 0.270  & 0.370  & 0.258 & 0.533 & 0.409 \\
    $\checkmark$     &       & $\checkmark$     & $\checkmark$     & 0.386 & 0.282 & 0.374 & 0.270  & 0.538 & 0.424 \\
    $\checkmark$     & $\checkmark$     &       & $\checkmark$     & 0.380  & 0.270  & 0.368 & 0.257 & 0.532 & 0.408 \\
    $\checkmark$     & $\checkmark$     & $\checkmark$     &       & 0.376 & 0.276 & 0.365 & 0.264 & 0.528 & 0.417 \\
    \midrule
    $\checkmark$     & $\checkmark$     & $\checkmark$     & $\checkmark$     & 0.398 & 0.293 & 0.387 & 0.281 & 0.552 & 0.437 \\
    \bottomrule
    \end{tabular}%
    }
  \label{tab:abilation2}%
\end{table}%

TABLE III and \ref{tab:abilation2} present the results of the ablation experiments on HKO-7 and ChinaNorth-2021, respectively. The model with only the spatiotemporal correlation layer serves as the baseline. In both tables,``T" stands for Temporal Modeling Layer, ``S" stands for Spatial Refinement Layer, ``ST" stands for Spatiotemporal Correlation Layer, and ``R" stands for Reconstruction module. In cells marked with a $``\checkmark"$, the corresponding module is employed, while blank cells indicate non-utilization of the corresponding module.

Table III presents the predictive performance of different module combinations on the HKO-7 dataset for various precipitation intensities. The baseline model, which includes only the Spatiotemporal Correlation Layer, exhibits relatively low CSI, GSS, and HSS across all precipitation intensities, indicating that considering only spatiotemporal correlation may not be sufficient to effectively capture the characteristics of different precipitation intensities. In the absence of the temporal modeling layer, the model exhibits a significant impact on the predictive performance for lower precipitation thresholds, which highlights the dependence of accurate modeling of the temporal evolution for forecasting weak precipitation. Similarly, without a spatial refinement layer, spatiotemporal correlation layer, and reconstruction module, the model's predictive performance also experiences a decline. TABLE \ref{tab:abilation2} presents the performance of different module combinations on the ChinaNorth2021 dataset for the mean (M in the table) and last time step (L in the table) predictions for the upcoming hour. The absence of the temporal modeling layer and spatiotemporal correlation layer results in a significant degradation of the performance metrics for the last time step, emphasizing the crucial role these two modules play in mitigating the long-term decay issue. The reconstruction module notably impacts the average performance over multiple extrapolation time steps. Without the spatial refinement module, there is also a decline in various metrics.

Overall, the ablation experiments demonstrate the importance of each component in the model. The temporal modeling layer, spatial refinement layer, and reconstruction module all contribute to improving the model's forecasting performance. The complete SFTformer model, encompassing all components, exhibits the best performance, confirming the effectiveness of the integrated approach.

\begin{figure}
\centering
\includegraphics[width=1.0\linewidth]{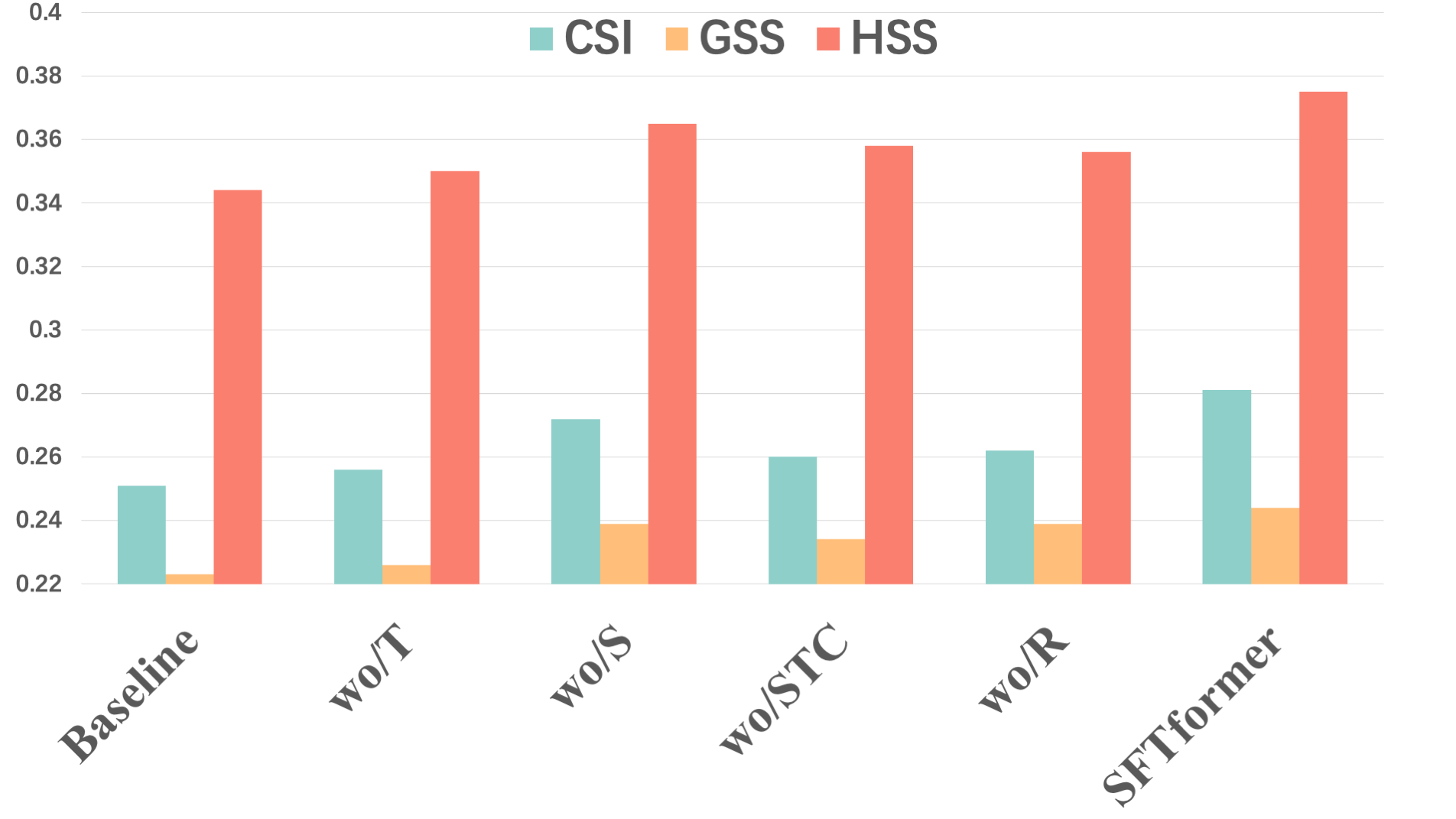}
\caption{The last-frame metrics of various ablation models on HKO-7.}
\label{fig_ablation1}
\end{figure}

\begin{figure}
\centering
\includegraphics[width=1.0\linewidth]{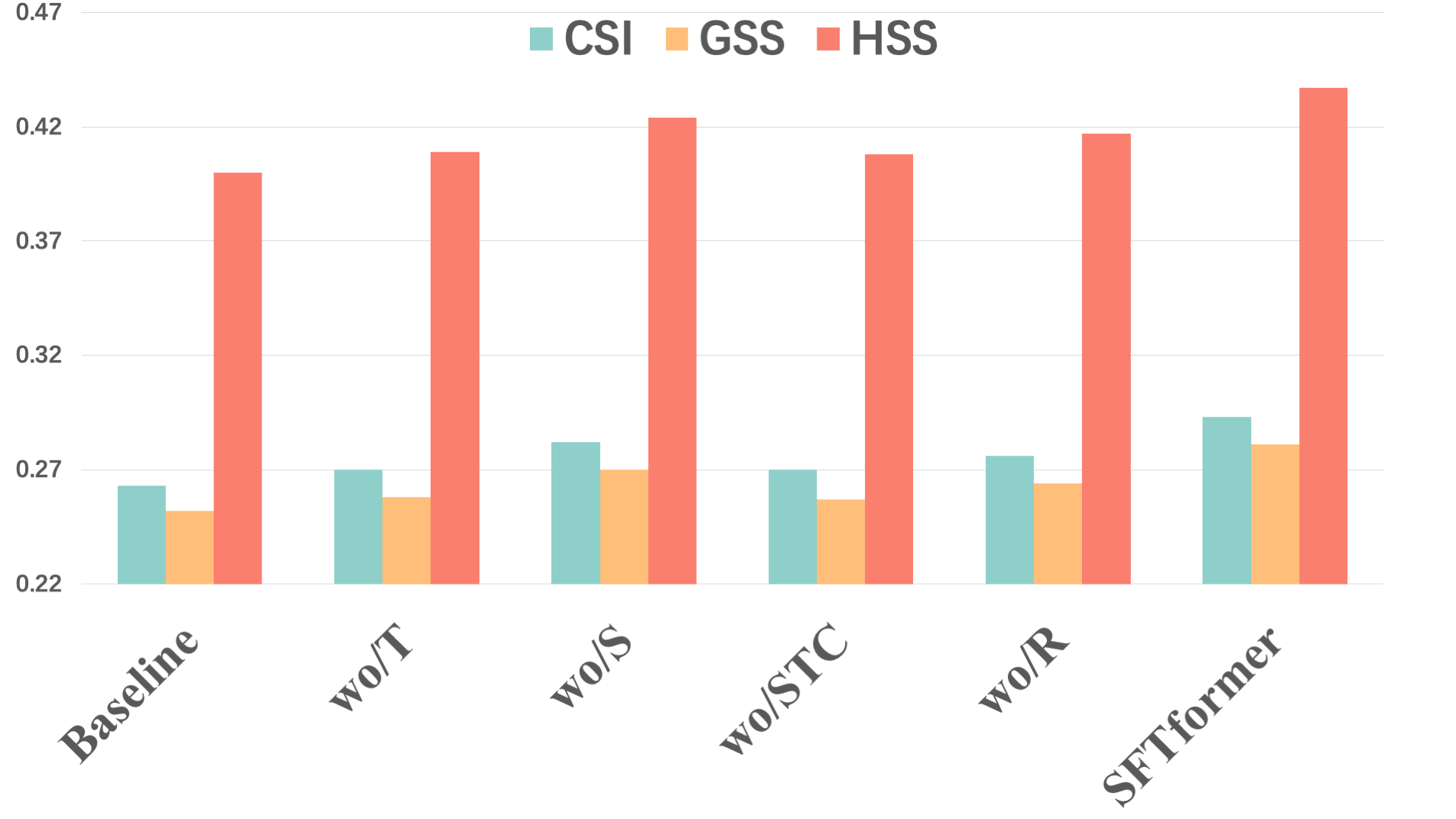}
\caption{The last-frame metrics of various ablation models on ChinaNorth-2021.}
\label{fig_ablation2}
\end{figure}

In order to visually demonstrate the contributions of each module to long-range forecasting, Fig. \ref{fig_ablation1} and Fig. \ref{fig_ablation2} present a comparison of various metrics for the last frame on HKO-7 and ChinaNorth-2021, respectively. The reconstruction module affects the prediction of the last frame, reaffirming that recalling information from historical frames is beneficial for forecasting future frames. Moreover, both the spatiotemporal correlation layer and the temporal modeling layer also exert notable effects on the prediction of the last frame. This is due to the fact that spatiotemporal feature correlations provide more prior knowledge for long-term forecasting, enabling the features to complement each other. Meanwhile, the temporal modeling layer not only ensures equal attention to historical information, preventing historical context from being forgotten but also leverages frequency analysis to grasp the cyclic pattern of echo evolution, significantly enhancing the model's long-term prediction capability. The spatial refinement layer has some impact on the last frame's results, though not as significant, which contributes to refining the details in radar echo morphology. The analysis above clearly emphasizes the impactful role of each module in long-term forecasting.
\section{Conclusion} \label{sec:conclusion}
In general, we propose a novel Transformer-based radar echo extrapolation framework called SFTformer. We design the SFT-Block to not only extract the correlation of spatiotemporal dynamics of radar echo but also decouple the temporal evolution from spatial morphology refinement, avoiding interference between them. The temporal modeling layer not only ensures equal attention to historical information, preventing historical context from being forgotten but also leverages frequency analysis to grasp the cyclic pattern of  echo evolution, significantly enhancing the model's long-term prediction capability. Furthermore, to strengthen the memory of historical information, a joint training paradigm for reconstructing historical echo sequences and predicting future echo sequences is proposed. By reconstructing historical echo features alongside the prediction task, the model is able to generate more accurate predictions. The proposed method is applied to the HKO-7 dataset and the ChinaNorth-2021 dataset, and the results demonstrate the superior performance of SFTformer in extrapolating future radar echoes for 1-hour, 2-hour, and 3-hour compared to other classical methods in the spatiotemporal prediction domain. Quantitative and qualitative analyses further confirm the effectiveness of SFTformer in precipitation forecasting. {However, there are still some issues to be addressed in future research. On one hand, further model simplification is needed to alleviate overfitting and achieve higher inference speed without compromising model performance. On the other hand, the consideration of introducing new architectures (such as GANs and diffusion models) or other constraints to generate future radar images that are clearer and closer to the ground truth will also be explored.}

% use section* for acknowledgment
\section*{Acknowledgment}
This work was supported by the National Key R\&D Program of China (2021YFB3900504).

	% Can use something like this to put references on a page
	% by themselves when using endfloat and the captionsoff option.
	\ifCLASSOPTIONcaptionsoff
	\newpage
	\fi

\bibliographystyle{ieeetr} 
\bibliography{main}

\end{document}